\documentclass[final,1p,times]{elsarticle}

\usepackage{amssymb}
\usepackage{lineno,hyperref}
\usepackage{xurl}
\usepackage{booktabs} 
\usepackage{tabularx} 
\newcommand{\tableheadline}[1]{\multicolumn{1}{l}{\textbf{#1}}}
\usepackage{amsthm}  
\usepackage{rotating}  
\usepackage{amsmath}
\usepackage{amsfonts}
\usepackage{comment}
\usepackage{algorithm}%
\usepackage{graphicx}
\usepackage{algpseudocode}
\usepackage{hyperref} 
\usepackage{soul}
\usepackage{mathrsfs}
\usepackage{enumitem}
\theoremstyle{definition}
\newtheorem{remark}{Remark}[section]
\newtheorem{definition}{Definition}
\newtheorem{proposition}{Proposition}
\newtheorem{corollary}{Corollary}
\newtheorem{lemma}{Lemma}
\DeclareMathOperator{\Nodes}{Nodes}

\DeclareMathOperator{\Var}{Var}

\journal{Applied Mathematical Modelling}

\begin{document}

\begin{frontmatter}



\title{A novel collision model for inextensible textiles and its experimental validation}


\author[mymainaddress]{Franco Coltraro\corref{mycorrespondingauthor}}
\cortext[mycorrespondingauthor]{Corresponding author}
\ead{franco.coltraro@upc.edu}

\author[mysecondaryaddress]{Jaume Amor\'os}
\author[mymainaddress]{Maria Alberich-Carramiñana}
\author[mymainaddress]{Carme Torras}

\address[mymainaddress]{Institut de Rob\`otica i Inform\`atica Industrial, CSIC-UPC, Barcelona, Spain.}
\address[mysecondaryaddress]{Universitat Polit\`ecnica de Catalunya, Barcelona, Spain.}

\begin{abstract}
	In this work, we introduce a collision model specifically tailored for the simulation of inextensible textiles. The model considers friction, contacts, and inextensibility constraints all at the same time without any decoupling. Self-collisions are modeled in a natural way that allows considering the thickness of cloth without introducing unwanted oscillations. The discretization of the equations of motion leads naturally to a sequence of quadratic problems with inequality and equality constraints. In order to solve these problems efficiently, we develop a novel active-set algorithm that takes into account past active constraints to accelerate the resolution of unresolved contacts. We put to a test the developed collision procedure with diverse scenarios involving static and dynamic friction, sharp objects, and complex-topology folding sequences. Finally, we perform an experimental validation of the collision model by comparing simulations with recordings of real textiles as given by a \textit{Motion Capture System}. The results are very accurate, having errors around 1 cm for DIN A2 textiles (42 x 59.4 cm) even in difficult scenarios involving fast and strong hits with a rigid object.
	
\end{abstract}

\begin{graphicalabstract}
	\begin{center}
		\includegraphics[width=1\linewidth]{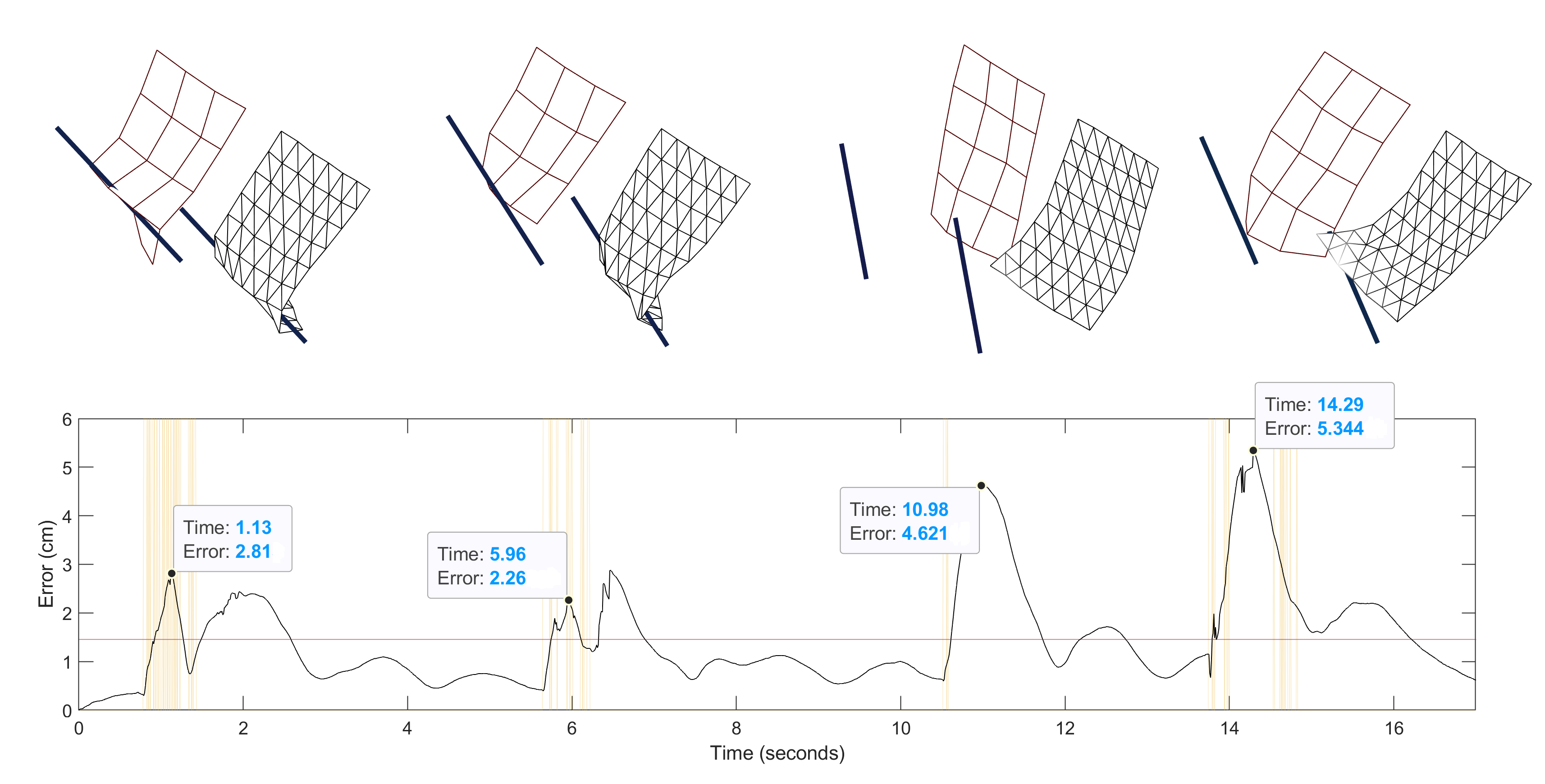} \label{palo_poliester} 
	\end{center}
	
	In this work, we present a novel collision model for inextensible textiles and validate it experimentally through a comparison with real textiles as recorded by a \textit{Motion Capture System}. In the figure we can see four frames comparing the recorded hitting with a stick of a DIN A2 (42 x 59.4 cm) polyester sheet (left) with its inextensible simulation (right); its average error being 1.44 cm. The recordings are obtained by attaching 20 reflective markers of diameter $3$ mm and weight of $0.013$ g to the real textile and following their trajectory. On the bottom, we show a full plot of the mean absolute error of the position of each marker with respect to its simulation and with yellow lines we highlight the moments in which the object is in contact with the cloth. Notice that the biggest errors appear not during the hits but just after because of aerodynamic effects. The above simulations, with a $7\times 9$ mesh, are two times faster than real-time. Our custom active-set solver is three times faster than a standard interior-point method using the same mesh resolution.
\end{graphicalabstract}


\begin{highlights}
	\item We present a novel collision model for inextensible cloth which results in a non-decoupled resolution of friction, inextensibility, and contact forces.
	
	\item Self-collisions are modeled in a way that allows considering the thickness of the cloth without introducing unwanted oscillations.
	
	\item An efficient active-set solver for the integration of the system is developed, which --as opposed to traditional methods--  can start from any non-necessarily feasible point. 
	
	\item We perform a successful empirical validation of the model, with scenarios involving low and high-friction surfaces and strong hits with a stick.
\end{highlights}

\begin{keyword}
	inextensible cloth \sep collision modeling \sep Coloumb friction \sep experimental validation \sep constrained optimization\\
	
	\MSC[2010] 70F35\sep 68U20 \sep 90C20
	
	
	
\end{keyword}

\end{frontmatter}


\section{Introduction and related work}

Textile objects are important and omnipresent in many relevant scenarios of everyday life, e.g. domestic, healthcare, or industrial contexts. However, as opposed to rigid objects, whose position in space is determined by position and orientation (i.e. six degrees of freedom), textile objects are challenging to handle in an automated fashion (e.g. by robots) because they change shape under contact and motion, resulting in an infinite-dimensional configuration space (when considered as continuous surfaces in 3D space). This huge dimensional jump makes existing perception and manipulation methods difficult to apply to textiles. Recent reviews on cloth manipulation, like \cite{sanchez2018robotic, yin2021modeling}, agree on the need to find simple, but realistic, models that enable more powerful learning methods to solve different problems related to cloth manipulation. 

\medskip

In \cite{COLTRARO2022} the authors introduced and discussed a physical model of cloth intended for its control under robotic manipulation in a human environment, which means that the textiles are subjected to moderate to low stresses \cite{Borras2020}. Because of its control purpose, the authors sought a model with which simulations of the motion of cloth could be computed fast, but with a small margin of error in the position of every point of the cloth when compared to its position in the real cloth subjected to the same manipulation by the robot. This led to the most basic hypothesis of the model, i.e. to assume that textiles are {\em inextensible}, that is, the surfaces that represent them only deform isometrically through space, aiming at preserving not only the area but also both dimensions of each piece of the cloth. This assumption, which simplifies the model by removing all considerations of elasticity in it, was shown to be very realistic for several materials and motions in \cite{COLTRARO2022}.

\medskip

Once we have a working inextensible cloth simulator that describes the internal dynamics of cloth; a related problem that always arises with inextensibility simulations is how to conciliate them with collision and friction forces (i.e. contacts with an obstacle or self-collisions), since both inextensibility and contact forces are hard constraints that the cloth must satisfy at the same time (moreover, friction forces depend on the magnitude of the contact force). In the conclusions of \cite{Goldenthal:2007:ESI}, it is stated: 

\medskip

\textit{"(...) there is no longer an efficient way to perfectly enforce both ideal inextensibility and ideal collision handling, since one filter must execute before the other, and both ideals correspond to sharp constraints. To enforce both perfectly would require combining them in a single pass, an elegant and exciting prospect from the standpoint of theory, but one which is likely to introduce considerable complexity and convergence challenges."}

\medskip

To our knowledge, the vast majority of models in literature decouple friction, contacts, and strain limiting (i.e. inextensibility), with all the possible artifacts that this introduces. In this work, we develop a novel collision model, its discretization and an active set-solver that can be seen as an extension of the \textit{fast projection algorithm} of \cite{Goldenthal:2007:ESI}, in order to incorporate contacts, friction and inextensibility in a single pass.

\subsection{Contributions}

In the following, we list the most important contributions of this article:

\begin{description}

\item[-] We present a novel collision model for cloth which results in a non-decoupled resolution of friction, strain limiting and contact constraints in a single pass.

\item[-] Self-collisions are modeled and detected in a way that allows considering the thickness of cloth without introducing unwanted oscillations.

\item[-] An efficient active-set solver for the integration of the system is developed, which --as opposed to traditional methods--  can start from any non-necessarily feasible point. 

\item[-] We present an empirical validation of the model, achieving errors of around 1 cm in challenging scenarios involving low and high-friction surfaces and high velocity collisions with a rigid object.

\end{description}

\subsection{Related Work}\label{sec_state_cols}
There is a rich history of research on contact and collisions for cloth simulation; in the following, we will review what we consider the most relevant methods, with an emphasis on newer works. We focus especially on articles that model collisions in physically different manners or use novel numerical algorithms to resolve them (and not so much on performance, e.g. GPU implementations of existing methods). Most of these works come from the Computer Graphics (CG) community and not so much from the Textile Engineering fields. This is due to the fact that in CG applications a core concern has always been simulating dressed moving mannequins (e.g. for movies and video games), whereas for textile engineers the focus has been on measuring intrinsic fabric properties. There are 3 main types of collision-response methods: 

\medskip

\noindent \textbf{Penalty-based methods:} these include very stiff spring-like forces of the form $kf(\epsilon)$ (where $\epsilon$ is the detected penetration depth) into the dynamical system when a penetration is detected. They are easy to implement and can work for simple cases, but are not physically accurate (e.g. they do not conserve momentum during the collision, see \cite{Smith2012ReflectionsOS}) and introduce a lot of stiffness into the system when $k$ is large (making it harder to integrate numerically). For example, Provot \cite{Provot1997CollisionAS} is one of the first to propose a penalty-based approach to solve collisions for cloth modeled as a mass-spring system. However, his method has no theoretical guarantees when there is more than one simultaneous collision. That is why, when many collisions accumulate during the same time step, he must resort to a fail-safe consisting of rigidifying zones of the cloth. On the other hand, he is among the first to give a formula to detect continuous-time collisions (i.e. when two moving edges or a node and triangle cross, see Section \ref{sec_autocols}). Finally, one of the main theoretical problems with penalty-based methods is that there is always a fast or strong enough collision where they fail because the spring force is not strong enough (although there are sophistications with more guarantees such as \cite{Harmon2009AsynchronousCM}). Despite their mentioned limitations, penalty methods are still widely used because of their ease of implementation (see, e.g. \cite{Geilinger2010ADD}).  

\medskip

\noindent \textbf{Impulse-based methods:} these methods include impulse forces (mostly based on rigid-body mechanical ideas) which are then used to modify velocities (and thus positions) instantaneously. They work well for individual collisions and are fast to compute, but run into problems for multiple simultaneous collisions. In this line, the work by Bridson et al. \cite{Bridson2005RobustTO} is considered to be by the CG community the first truly robust method for handling collisions, contact and friction for cloth simulation. More than a unified physical model, their method consists of a list of procedures used to get a state of cloth that is collision-free but not necessarily physically realistic. They first apply penalty forces as a prevention method and then impulse forces for detected collisions in continuous time. When there are simultaneous collisions in the same time step, after a fixed number of iterations they also resort to rigid impact zones (but with corrected formulas with respect to \cite{Provot1997CollisionAS}). Some alternatives exist to avoid the rigidification or areas of the cloth, e.g. \cite{Volino2000AccurateCR} and \cite{Harmon2008RobustTO}. These two methods derive impulse forces for the case of simultaneous collisions with the aid of constraints that are being violated by the detected penetrations. Their main problem is that in order to be efficient they derive the impulses from equality constraints, and this introduces \textit{sticking} artifacts into the simulation (some nodes are forced to stay in contact, when they otherwise would depart). Moreover, another issue with impulse-based methods is that since one is modifying positions instantaneously, strain-limiting procedures must be performed prior to (or after) collision response, and thus constraints such as inextensibility cannot be maintained exactly. These type of models have somehow fallen out of fashion in recent times (at least in the research literature) in favor of constraint-based approaches.

\medskip

\noindent \textbf{Constraint-based methods:} with the increase of computing power in the last decades these methods have flourished from a research viewpoint. Their idea is simple: once a collision is detected, a constraint is defined (which is being violated because of the collision) and an optimization problem should be solved with all the detected constraints. Most methods vary mainly in how the optimization problem is solved and how friction is modeled (going from exact Coloumb models to linearized ones). These restrictions can be imposed as equalities or inequalities. As already said, imposing the constraints as equalities can be very efficient but one runs into \textit{sticking artifacts} (since some constraints can pull from others) and thus it is better to consider them inequalities. This is in turn known as the Signorini-contact model. Otaduy et al. \cite{Otaduy2009ImplicitCH} were among the first to propose a physically sound constrained dynamics formulation for cloth simulation and contact. They employ Signorini’s contact model and add to it a linearized Coulomb's friction model. The optimization program is stated formally as a linear complementary problem (LCP) and friction and contacts are afterwards decoupled in order to be solved numerically. 

\smallskip

Years later, Li et al. \cite{Li2018AnIF} implemented exact Coulomb friction for cloth simulation using adaptive meshes. Their constraint-based solver (released later as an open-source simulator called ARGUS) is costly to run but treats contacts (and friction) simultaneously and implicitly. Recently, Ly et al. \cite{Ly2020ProjectiveDW} proposed an alternative numerical algorithm based on Projective Dynamics that accelerates by an order of magnitude results obtained with ARGUS (see also \cite{li2022diffcloth} for a differentiable cloth model based on Projective Dynamics). Their main drawback is that they inherit the limitations of Projective Dynamics, in particular, the lack of a simple rule to ensure convergence. Moreover, they do not consider strain limiting and collisions simultaneously and hence depict cloth as a very elastic material. One of the main limitations of all the previous methods is the difficulty of integrating at the same time strain limiting (e.g. inextensibility) with the collision handling algorithm. In this line, Li et al. \cite{Li2021CodimensionalIP} have developed a method that integrates strain limiting and collisions in a single pass with the extensive use of barrier functions. They also propose a benchmark set of challenging tests. Finally, we mention that, naturally, constraint-based methods also have some drawbacks: most of the time they need dedicated solvers and thus can be cumbersome to implement and since several optimization problems must be solved, they are in general slower than impulse or penalty-based methods.

\subsection{Overview}

The model we will derive in this work lies in the category of constraint-based methods and hence can handle efficiently simultaneous collisions. We will solve a quadratic problem with inequality constraints and therefore we will be employing the physically accurate model of Signorini. In order to do so, we will develop a novel active-set solver to resolve collisions efficiently. Moreover, we will derive a simple friction model that allows us to integrate all forces and constraints in a simple pass without the need to decouple contact and friction forces like it has been traditionally done for rigid body contacts (see \cite{Kaufman2008StaggeredPF}). Finally, our method considers strain limiting (inextensibility) and contact at the same time, unlike most current methods. Our algorithm can be seen as an extension of the \textit{fast projection algorithm} (see \cite{Goldenthal:2007:ESI}) developed in order to incorporate contacts, friction and inextensibility in a single pass. 

\subsection{Organization}

The rest of this paper is organized as follows: in Section \ref{sec_cols_fric} we explain how to introduce contacts and friction into the equations of motion in a physically sound manner. Next, in Section \ref{sec_autocols} we delve into the problem of detecting self-collisions and resolving them, including how to take into account the thickness of the cloth during this process. In Section \ref{sec_discr_qua} we explain how to discretize the model in order to integrate it numerically without decoupling contacts, friction and inextensibility constraints, as it is usually done. In Section \ref{sec_active-set} we present a novel algorithm to solve efficiently the quadratic problems that arise from the previous discretization. Finally in Sections \ref{sec_eval_cols} and \ref{sec_expvalidation} we evaluate the presented model, including the simulation of challenging scenarios and performing an empirical validation of the collision model, by comparing it with real data.

\section{Modeling of contacts and friction} \label{sec_cols_fric}
Assuming that $S$ is an inextensible surface moving through space; in \cite{COLTRARO2022} the authors derived the following time- and space-dependent constraints 
\begin{equation}\label{constraints}
	\langle\varphi_{\xi},\varphi_{\xi}\rangle(t) = E_0,\; \langle\varphi_{\xi},\varphi_{\eta}\rangle(t) = F_0,\; \langle\varphi_{\eta},\varphi_{\eta}\rangle(t) = G_0\; \text{for }t\geq 0,
\end{equation}
where $\varphi(\xi,\eta)$ is any smooth parametrization of $S$, $t\geq 0$ represents time, $E_0,F_0,G_0$ are the coefficients of the first fundamental form of $S$ and are constant in time (but not necessarily in space), and $\varphi_{\xi} = \partial_{\xi}\varphi$ denotes the partial derivative with respect to $\xi$. In that same work, it was explained how to efficiently discretize this system of partial differential equations with the aid of finite elements, and how to include them in the equations of motion of a textile in order to model realistically the dynamics of inextensible cloth.

\remark[Notation] For the remaining of this work we will assume that the cloth $S$ has been discretized into a triangular or quadrilateral mesh and the position of its $N$ vertices is given by $\boldsymbol{\varphi}(t) = (\textbf{x}(t),\textbf{y}(t),\textbf{z}(t))^\intercal\in\mathbb{R}^{3N}$. Moreover, when integrating numerically the equations of motion of the cloth (see Equation (\ref{eq_colisiones})), as usual, we will approximate $ \boldsymbol{\varphi}(t)$ and $\dot{ \boldsymbol{\varphi}}(t)$ by $\{ \boldsymbol{\varphi}^0, \boldsymbol{\varphi}^1,\dots\}$ and $\{\dot{ \boldsymbol{\varphi}}^0,\dot{ \boldsymbol{\varphi}}^1,\dots\}$, where $ \boldsymbol{\varphi}^n$ and $\dot{ \boldsymbol{\varphi}}^n$ are the position and velocities of the nodes of the mesh at time $t_n = n\cdot dt$ and $dt > 0$ is the size of the chosen time step.

\medskip

For its application in the real world, we also need to include in our model collisions of the cloth with an object (e.g. a table) and with itself. We will model this by enforcing a set of constraints $\textbf{H}( \boldsymbol{\varphi}) \geq 0$, which we assume have a well-defined outwards normal $\nabla\textbf{H}( \boldsymbol{\varphi})$ (almost everywhere). Observe that the obstacle could move in time, but we need to know its position. We can then model collisions by including new (non-smooth, see \cite{Kun00} and \cite{Zho93}) forces into the equations of motion. Signorini's contact model then reads (see \cite{Jean1999TheNC}):
\begin{equation}\label{eq_colisiones}
\begin{cases}
	\textbf{M}\ddot{ \boldsymbol{\varphi}} = \textbf{F}( \boldsymbol{\varphi},\dot{ \boldsymbol{\varphi}}) - \nabla\textbf{C}( \boldsymbol{\varphi})^\intercal \boldsymbol{\lambda} + \nabla\textbf{H}( \boldsymbol{\varphi})^\intercal \boldsymbol{\gamma},\\
	\textbf{C}( \boldsymbol{\varphi}) = 0,\\
	\textbf{H}( \boldsymbol{\varphi}) \geq 0,\quad  \boldsymbol{\gamma}\geq 0,\quad  \boldsymbol{\gamma}^\intercal\cdot \textbf{H}(\varphi) = 0,
\end{cases}
\end{equation}
where $\textbf{M}$ is the mass matrix (including the cloth's density),  $\textbf{C}( \boldsymbol{\varphi}) = 0$ are the discretization of the inextensibility constraints (\ref{constraints}) and $\boldsymbol{\lambda}$ its associated Lagrange multipliers, $\boldsymbol{\gamma}\geq 0$ are new contact Lagrange multipliers and we have grouped in the force term $\textbf{F}( \boldsymbol{\varphi},\dot{ \boldsymbol{\varphi}})$ damping, bending, gravity and aerodynamic forces. Now the system is non-smooth, which is why we will need to use a first-order (implicit) integration scheme \cite{Kun00}. A simple model for friction can be introduced if we  add yet another force of the form: 

\begin{equation}
\textbf{f}_{\mu}(\dot{ \boldsymbol{\varphi}}) = -\mu\textbf{V}(\dot{ \boldsymbol{\varphi}})^\intercal\boldsymbol{\beta}
\end{equation}
where $\mu>0$ is a friction constant, $\boldsymbol{\beta}$ are new multipliers (one for each contact constraint) satisfying that they belong to the friction's cone, i.e. they satisfy component-wise  $\beta_i\leq||\nabla H_i(\boldsymbol{\varphi})^\intercal\gamma_i||$, and $\textbf{V}(\dot{\boldsymbol{\varphi}})$ are unit (relative) tangent velocities at the points of contact, i.e. for the case of a collision with a static obstacle:
\begin{equation*}
kV_i(\dot{\boldsymbol{\varphi}}) = \dot{\boldsymbol{\varphi}} - \langle\dot{\boldsymbol{\varphi}},\textbf{n}_i\rangle\cdot \textbf{n}_i,
\end{equation*}
where $\textbf{n}_i = \frac{\nabla H_i(\boldsymbol{\varphi})}{||\nabla H_i(\boldsymbol{\varphi})||}$ and $k$ is a normalization constant. For theoretical details and more sophisticated models for friction see \cite{Acary2008NumericalMF}.

\remark We now list some implicit assumptions we are making in stating the collision model as Equation (\ref{eq_colisiones}):

\begin{enumerate}
\item When $H = 0$ defines a surface (e.g. a plane or a sphere), the condition $\textbf{H}( \boldsymbol{\varphi}) \geq 0$ means that for each node $p_i$ of the cloth's mesh we impose
\begin{equation*}
	H_i(\boldsymbol{\varphi}) := H(p_i(t)) \geq 0.
\end{equation*}
This only forces the vertices of the cloth to be outside the obstacle (but there could be some penetrations of the faces). When the mesh is fine enough this is not really a problem, in the case of coarse meshes one can add yet another constraint for the middle point of each face.

\item Signorini's condition implies that when there is no contact taking place, i.e. $H_i( \boldsymbol{\varphi}) > 0$, then there is no repulsive force acting, i.e. $\nabla H_i(\boldsymbol{\varphi})^\intercal\gamma_i = 0$. Therefore there is also no friction force acting, i.e. $\beta_i = 0$.

\item Without any other condition the multipliers $\boldsymbol{\beta}$ are not uniquely defined. A common approach is to assume that these multipliers cause \textit{maximal dissipation} (see \cite{Jean1999TheNC,Kaufman2008StaggeredPF,Smith2012ReflectionsOS}). This amounts to solving a linear program. In practice, we will assume that $\beta_i = ||\nabla H_i(\boldsymbol{\varphi})^\intercal\gamma_i||$ (which is anyways always the case when the tangent velocity is nonzero).

\item This model assumes that the collision is inelastic (there is no bouncing). This is a reasonable assumption for cloth; we will corroborate this in Section \ref{sec_expvalidation} when we perform the empirical validation of the collision model. 
\end{enumerate} 

\section{Modeling and detection of self-collisions}\label{sec_autocols}
In this section, we explain how to detect and include self-collision constraints under the framework presented in Section \ref{sec_cols_fric}. Particularly important for efficiency and to avoid unwanted oscillations is how to take into account the thickness of cloth. 

\medskip

The goal is to define the constraints $H_k$ that account for modeling self-collisions of the cloth inside the function $\textbf{H}(\boldsymbol{\varphi})\geq 0$. In principle we need to integrate numerically the equations of motion (\ref{eq_colisiones}) and advance the simulation from $\boldsymbol{\varphi}^{n}$ to $\boldsymbol{\varphi}^{n+1}$, then check if in the process self-collisions took place, and in case they did, add new constraints $H_k$ to the system and repeat the numerical integration. This process must be repeated until no new collisions are found. In practice this is costly and thus we will develop a more efficient procedure that takes advantage of the way we integrate numerically the equations of motion. For the time being, assume we have both $\boldsymbol{\varphi}^{n}$ and $\boldsymbol{\varphi}^{n+1}$ (and their velocities) available to make computations. 

\subsection{Detection of self-collisions}\label{detection}
In general, we assume that the cloth is triangulated (in case of a quadrangulation we can always divide the quads in two); then in case of collision, there are only two stable (i.e. detectable) possibilities: an edge-edge collision and a node-face collision. In these two cases, we have four nodes involved which at some instant of time belong to the same plane. We must then only check if two co-planar segments cross or if a point is within a triangle. These two problems are readily solved using barycentric coordinates. 

\medskip

Now we describe in more detail the process: in order to save computational time, we only check if a collision has happened for pairs of edges (or nodes and faces) that at time $t_n$ or $t_{n+1}$ are \textit{sufficiently} close. To obtain this list of sufficiently close (up to some tolerance) pairs, a \textit{hierarchical method} is used (see e.g.\cite{Provot1997CollisionAS}). Next, denoting by $x_1,x_2,x_3,x_4$ the position of the four \textit{candidate} nodes at time $t_n$ and by $v_1,v_2,v_3,v_4$ their velocities at $t_{n+1}$, we must check if for some $t$:

\begin{equation}\label{autocol_det}
\det(\tilde{x}_1 + t\cdot\tilde{v}_1,\tilde{x}_2 + t\cdot\tilde{v}_2,\tilde{x}_3 + t\cdot\tilde{v}_3) = 0
\end{equation}
where $\tilde{x}_i = x_i - x_4$ and $\tilde{v}_i = v_i - v_4$, since $\boldsymbol{\varphi}^{n+1} = \boldsymbol{\varphi}^{n} + dt\cdot\dot{\boldsymbol{\varphi}}^{n+1}$. This is a cubic equation $a_3t^3 + a_2t^2 + a_1t + a_0 = 0$ whose coefficients are easily computed by expanding the previous determinant.

\medskip

When $t\ll dt$ is small, the solution of the previous equation can be approximated linearly by $-\tfrac{a_0}{a_1}$. In any case, if there is a root for some $t_c\in[0,dt]$, we must then do two different calculations with the four co-planar points $y_i = x_i + t_c\cdot v_i$ in order to see if a collision has occurred. Namely: in the edge-edge case we check if the two co-planar segments intersect and in the node-face case check if the node is inside the (triangular) face. This is a trivial plane geometry problem solved by using barycentric coordinates.

\subsection{Constraint definition for self-collisions}\label{response}

Once we have detected a self-collision, we now describe the computation of the response constraint $H_k$. It will be linear in $\boldsymbol{\varphi}$ and naturally have slightly different forms depending on our two cases:

\begin{enumerate}
\item Edge-edge case: 
\begin{equation*}
	H_k(\boldsymbol{\varphi}) :=\langle \pi_{\alpha}(x_1,x_2) - \pi_{\beta}(x_3,x_4) ,\nu\rangle \geq 0,
\end{equation*}
where $x_i$ are the four endpoints of the two edges, $\pi_{\alpha}(x_1,x_2) = (1-\alpha)x_1 +\alpha x_2$ and $\pi_{\beta}(x_3,x_4) = (1-\beta)x_3 +\beta x_4$ are the closest points between the two segments and $\nu$ is the normal vector to both edges. In general, the values $\nu,\alpha,\beta$ vary with time. We will nevertheless assume that they are constant during the time-step, and compute them with the positions of the segments given by  $\boldsymbol{\varphi}^{n+1}$. The normal vector $\nu$ is oriented such that $H_k(\boldsymbol{\varphi}^{n})\geq 0$.

\item Node-face case: 
\begin{equation*}
	H_k(\boldsymbol{\varphi}) :=\langle x_4 - \pi(x_1,x_2,x_3) ,\nu\rangle  \geq 0,
\end{equation*}
where $x_4$ is the node, $x_i$ are the three corners of the triangle, again $\pi(x_1,x_2,x_3) = ux_1 + vx_2 + wx_3$ is the closest point inside the face to the node and $\nu$ is the normal vector to the triangle. In general, the values $\nu,u,v,w$ vary with time. We will again assume that they are constant in time, and compute them with the positions given by $\boldsymbol{\varphi}^{n+1}$. The normal vector $\nu$ is oriented such that $H_k(\boldsymbol{\varphi}^{n})\geq 0$.
\end{enumerate}

\remark Notice that:

\begin{enumerate}
\item By construction $H_k(\boldsymbol{\varphi}^{n+1}) < 0$.
\item The constraint $H_k$ is an approximation of the signed distance between the pairs edge-edge and node-face (only an approximation since $\nu$ and the barycentric coefficients are fixed in time).
\item Since in practice cloth has thickness, say $\tau_0$, the constraint we actually must impose is $H_k(\boldsymbol{\varphi})\geq\tau_0$.
\end{enumerate}

\subsection{Proximity constraints and cloth thickness}\label{grosor}

Adding the constraints we have just defined is enough to correct all present self-intersections. Nevertheless, there are two main drawbacks:

\begin{enumerate}
\item Efficiency: most cloth self-intersections can be avoided before they happen by adding preventive constraints. 
\item Vibrations: since we are assuming that the cloth has a thickness $\tau_0 > 0$, when we integrate the system again and go from $H_k(\boldsymbol{\varphi}^{n}) < 0$ to $H_k(\boldsymbol{\varphi}^{n+1}) \geq \tau_0$, the change between the position of the nodes can be too large, and since our cloth is inextensible, this could create unwanted oscillations. 
\end{enumerate} 

In order to avoid these two problems, we apply the detection procedure previously explained in \ref{detection} with one small difference: during the detection phase we move the pairs (edge-edge or face-node) closer, using their normal vectors and taking into account the thickness of the cloth, so that pairs that are too close and/or are approaching each other, are kept at a minimum distance of $\tau_0$ before they actually cross. Since the restrictions we are considering are inequalities, we can add these to the system because they only affect the dynamics of cloth in case the constraint will actually get violated. In symbols, this means that we compute  Equation (\ref{autocol_det}) of the third degree polynomial using the altered positions given by $\hat{x}_i = x_i \pm \omega\tau_0 \nu$, where $\nu$ is the unit normal vector (the cross product for the edge-edge case and the normal to the triangle for node-face case), $\omega\approx 0.5$ is what we will call a \textit{proximity parameter} and the sign $\pm$ is chosen so that the pairs approach each other. Afterwards the response constraint $H_k$ is calculated as usual (i.e. the normals and the barycentric coordinates) with the unaltered positions $x_i$ given by $\boldsymbol{\varphi}^{n+1}$. 

\medskip

Although there are algorithms that estimate time-to-contact based on trajectories, update these estimates with a frequency dependent on the velocities of the involved pairs, and perform collision tests only for segments that are close to contact taking into account the thickness of the simulated material, we found that these methods did not perform well when considering inextensible cloth, and added unwanted oscillation not present when using the method previously described using the proximity parameter $\omega$.

\remark \label{obs_grosor} It is usually enough to use the positions $\hat{x}_i = x_i \pm\omega\tau_0\nu$, where $\omega\approx 0.5$ to detect all self-collisions, nevertheless some can sometimes be missed because the nodes have moved too much. In that case, we enter an iterative process reducing gradually the value of $\omega$ until all are resolved. We will explain this in more detail in Section \ref{autocol_iter}.

\begin{definition}(Self-collision constraints). 
We will denote by 
\begin{equation*}
	\mathcal{C} = \text{Collisions}_{\omega}\left(\boldsymbol{\varphi}^{n} \rightarrow_{dt}\boldsymbol{\varphi}^{n+1}\right)
\end{equation*}
the set of self-collisions constraints detected when moving from the state $\boldsymbol{\varphi}^{n}$ to the state $\boldsymbol{\varphi}^{n+1}$ with proximity parameter $\frac{1}{2} > \omega \geq 0$.
\end{definition}

\section{Numerical integration of the system}\label{sec_discr_qua}
In this section, a novel numerical discretization is presented in order to integrate implicitly the extended equations of motion. This discretization leads naturally to a sequence of quadratic problems with inequality constraints. We explain in detail how to include self-collision constraints under this scheme.

\medskip

The friction force and the contact constraints introduced in Section \ref{sec_cols_fric} are in general highly non-linear and stiff and thus must be integrated implicitly. To integrate the system numerically from time $t_n$ to $t_{n+1}$ (i.e. to advance the simulation from $\boldsymbol{\varphi}^{n}$ to $\boldsymbol{\varphi}^{n+1}$), we perform as in \cite{COLTRARO2022} an iterative process  $\boldsymbol{\varphi}_{j+1} =  \boldsymbol{\varphi}_j + \Delta \boldsymbol{\varphi}_{j+1}$ where the initial point is the unconstrained step $ \boldsymbol{\varphi}_0 =  \boldsymbol{\varphi}_0^{n+1}(\boldsymbol{\varphi}^{n},\dot{\boldsymbol{\varphi}}^{n})$ given by an implicit Euler scheme. Also, we write:
\begin{equation*}
\textbf{H}( \boldsymbol{\varphi}_{j+1})= \textbf{H}( \boldsymbol{\varphi}_j + \Delta \boldsymbol{\varphi}_{j+1})\simeq \textbf{H}( \boldsymbol{\varphi}_{j}) + \nabla\textbf{H}( \boldsymbol{\varphi}_{j})\Delta \boldsymbol{\varphi}_{j+1},
\end{equation*}
and similarly 
\begin{equation*}
\textbf{C}( \boldsymbol{\varphi}_{j+1})= \textbf{C}( \boldsymbol{\varphi}_j + \Delta \boldsymbol{\varphi}_{j+1})\simeq \textbf{C}( \boldsymbol{\varphi}_{j}) + \nabla\textbf{C}( \boldsymbol{\varphi}_{j})\Delta \boldsymbol{\varphi}_{j+1},
\end{equation*}
and then solve iteratively the following sequence of quadratic programs with linear equality and inequality constraints:
\begin{equation}\label{prob_qua}
\begin{cases}
	\min_{\Delta \boldsymbol{\varphi}_{j+1}}\tfrac{1}{2}\Delta \boldsymbol{\varphi}_{j+1}^\intercal\cdot\textbf{M}\cdot\Delta \boldsymbol{\varphi}_{j+1} - \Delta \boldsymbol{\varphi}_{j+1}^\intercal\cdot\textbf{f}_{\mu}(\dot{ \boldsymbol{\varphi}}_j)  \\
	\textbf{C}( \boldsymbol{\varphi}_{j}) + \nabla\textbf{C}( \boldsymbol{\varphi}_{j})\Delta \boldsymbol{\varphi}_{j+1} = 0,\\
	\textbf{H}( \boldsymbol{\varphi}_{j}) + \nabla\textbf{H}( \boldsymbol{\varphi}_{j})\Delta \boldsymbol{\varphi}_{j+1} \geq 0,
\end{cases}
\end{equation}
where 

\begin{enumerate}
\item $\dot{\boldsymbol{\varphi}}_{j+1}=\frac{\boldsymbol{\varphi}_{j+1}-\boldsymbol{\varphi}^{n}}{dt}$ is an approximation of $\dot{\boldsymbol{\varphi}}^{n+1}$,
\item $\textbf{f}_{\mu}(\dot{ \boldsymbol{\varphi}}_j) = -\mu\textbf{V}(\dot{ \boldsymbol{\varphi}}_j)^\intercal\Delta\boldsymbol{\beta}_j$ is the friction force at iteration $j$,
\item $\textbf{V}(\dot{ \boldsymbol{\varphi}}_j)$ are the relative unit tangent velocities,
\item $(\Delta\beta_j)_i = ||\nabla H_i(\boldsymbol{\varphi}_j)^\intercal(\Delta\gamma_j)_i||$ is the magnitude of the contact forces at iteration $j$,
\item and $\Delta\boldsymbol{\gamma}_j\geq 0$ are the multipliers associated to the contact constraints.
\end{enumerate}
We iterate until
\begin{equation}\label{stop_crit}
\max|\textbf{C}(\boldsymbol{\varphi}_j)| < \epsilon_0,\quad \min\textbf{H}(\boldsymbol{\varphi}_j) \geq -\epsilon_1,\quad\max|\Delta\boldsymbol{\varphi}_j| < \epsilon_2
\end{equation}
for some tolerances $\epsilon_0,\epsilon_1,\epsilon_2 > 0$. This third condition ensures that the friction force has stabilized. Note that the critical points of the previous quadratic problems (\ref{prob_qua}) are:

\begin{equation} \label{KKTcoli}
\begin{cases}
	\textbf{M}\cdot\Delta \boldsymbol{ \boldsymbol{\varphi}}_{j+1} = -\nabla\textbf{C}( \boldsymbol{\varphi}_{j})^\intercal\Delta \boldsymbol{\lambda}_{j+1} + \nabla\textbf{H}( \boldsymbol{\varphi}_{j})^\intercal\Delta \boldsymbol{\gamma}_{j+1} -\mu\textbf{V}(\dot{ \boldsymbol{\varphi}}_j)^\intercal\Delta\boldsymbol{\beta}_j,\\
	\textbf{C}( \boldsymbol{\varphi}_{j}) + \nabla\textbf{C}( \boldsymbol{\varphi}_{j})\Delta \boldsymbol{\varphi}_{j+1} = 0,\\
	\textbf{H}( \boldsymbol{\varphi}_{j}) + \nabla\textbf{H}( \boldsymbol{\varphi}_{j})\Delta \boldsymbol{\varphi}_{j+1} \geq 0,\\ 
	\Delta \boldsymbol{\gamma}_{j+1}\geq 0,\quad \Delta \boldsymbol{\gamma}_{j+1}^\intercal\cdot \left[\textbf{H}( \boldsymbol{\varphi}_{j}) + \nabla\textbf{H}( \boldsymbol{\varphi}_{j})\Delta \boldsymbol{\varphi}_{j+1}\right] = 0,\\
	(\Delta\beta_j)_i = ||\nabla H_i(\boldsymbol{\varphi}_j)^\intercal(\Delta\gamma_j)_i||.
\end{cases}
\end{equation}

\remark In order to integrate friction force we have made the approximation $\textbf{f}_{\mu}(\dot{ \boldsymbol{\varphi}}_{j+1}) \simeq \textbf{f}_{\mu}(\dot{ \boldsymbol{\varphi}}_j)$. That is, we have dropped the gradient we would normally have with a first-order approximation (this is what we also do with the gradient of the constraint forces in the first equation of (\ref{KKTcoli}), see \cite{Goldenthal:2007:ESI} for more details).

\subsection{Addition of self-collision constraints}\label{autocol_iter}
Instead of checking and generating all self-collision constraints only with the states $\mathcal{C} = \text{Collisions}_{\omega}\left(\boldsymbol{\varphi}^{n} \rightarrow_{dt}\boldsymbol{\varphi}^{n+1}\right)$, we take advantage of the fact that we perform an iterative process. We now explain how we introduce self-collisions into the sequence of problems (\ref{prob_qua}) for every step. For every iteration $j$ we check for self-collisions (see Section \ref{detection}) taking into account the thickness of the cloth (Section \ref{grosor}) between the states $\boldsymbol{\varphi}^{n}$ and $\boldsymbol{\varphi}_{j}$ and generate the corresponding constraints (Section \ref{response}). In symbols this means that all the constraints $\mathcal{C}_j = \text{Collisions}_{\omega}\left(\boldsymbol{\varphi}^{n} \rightarrow_{dt}\boldsymbol{\varphi}_{j}\right)$ for $j\geq 0$ and $\omega\approx 0.5$ are added to the system. In the rare case that the same collision is found in two different iterations we only keep the constraint defined by the later iteration. Then, when we find a state $\boldsymbol{\varphi}_{j^*}$ that satisfies the stopping criteria (\ref{stop_crit}), we check for self-collisions with $\omega = 0$ between the states $\boldsymbol{\varphi}^{n}$ and $\boldsymbol{\varphi}_{j^*}$, and in case no self-collision is detected, we put $\boldsymbol{\varphi}^{n+1} = \boldsymbol{\varphi}_{j^*}$. Otherwise, we repeat the whole iteration process with a smaller value of $\omega$ (see Remark \ref{obs_grosor}).

\section{Efficient solution of the quadratic problems}\label{sec_active-set}

In this section, we study how to solve efficiently the sequence of quadratic problems defined before. We present a novel \textit{active-set} method tailored to our problem. A detailed procedure is laid out in pseudo-code in Algorithm \ref{algo_coli}.

\begin{definition}[Active constraint]
In a constrained optimization problem (such as (\ref{prob_qua})), we say that an inequality constraint $g(x)\geq0$ is active at a feasible point $y$ if $g(y) = 0$. Note that all equality (in our case inextensibility) constraints are always active.
\end{definition}
In order to solve the sequence of problems (\ref{prob_qua}) we could employ any quadratic problem solver, but we would not be taking advantage of the structure of our problem. That is, if in one of the iterations $j$ one of the contact constraints $H_i$ is active (see the previous definition), then it is likely that it will be active again at the next iteration. Physically, this means that nodes of the cloth that are in contact with an obstacle (or among themselves) at some iteration, are likely to remain in contact. This suggests the use of active-set-methods \cite{Nocedal1999NumericalO} to solve the quadratic problems. We will develop a novel \textit{active-set} algorithm in the following pages. Although we could use one of the many existing ones, they always require that one begins with a feasible (albeit not optimal) solution to the problem. Our method will not have this requirement.

\medskip

The main idea of active set methods is to find the \textit{active set} of constraints at the solution, because then, once known, the program can be solved by ignoring inactive constraints, and assuming that all active inequality constraints are equality constraints. Recall that solving quadratic problems with equality constraints is very cheap and can be done by solving a linear system (see \cite{Goldenthal:2007:ESI}). This will be precisely what we will do for every iteration of the sequence (\ref{prob_qua}). In order to find the active set, one splits the constraints in two sets: 

\begin{description}
\item \textit{The working set,} $\mathcal{W}$: these are the constraints believed to be active ($g = 0$) and therefore are imposed as equality constraints when one solves the optimization problem. This can be initialized as the set consisting only of equality constraints. 
\item \textit{The observation set,} $\mathcal{O}$: these are the constraints believed to be inactive ($g > 0$) and therefore are not imposed as equality constraints. Since they are not included in the problem one must be careful that they do not become violated. 
\end{description}

Then one proceeds as follows:

\begin{enumerate}
\item[1)] solve the equality problem defined by the working set;
\item[2)] compute the Lagrange multipliers of the working set for the inequality constraints;
\item[3)] send some subset of the constraints with negative Lagrange multipliers to the observation set;
\item[4)] if all multipliers are positive, check if all constraints in the observation set remain feasible;
\item[5)] send some subset of the infeasible constraints to the working set;
\item[6)] repeat.
\end{enumerate}

Then, if at some iteration we have found an increment $\Delta \boldsymbol{\varphi}_{j+1}$ such that all contact constraints in the working set have positive Lagrange multipliers $\Delta \boldsymbol{\gamma}_{j+1}\geq 0$ (see Equation (\ref{KKTcoli})) and all constraints in the observation set are not violated, we have found the active set (see \cite{Nocedal1999NumericalO}) and we can make the update $\boldsymbol{\varphi}_{j+1} =  \boldsymbol{\varphi}_j + \Delta \boldsymbol{\varphi}_{j+1}$. The following proposition ensures that we do not enter a never-ending cycle:

\begin{proposition}[Entry and exit of constraints]
Given the constrained linear system (with unknowns $\Delta\boldsymbol{\varphi}$)
\begin{equation}
	\begin{cases}
		\textbf{M}\Delta\boldsymbol{\varphi} = \nabla\textbf{H}( \boldsymbol{\varphi})^\intercal \Delta\boldsymbol{\gamma},\\
		\textbf{H}( \boldsymbol{\varphi}) + \nabla\textbf{H}( \boldsymbol{\varphi})\Delta \boldsymbol{\varphi} = 0,
	\end{cases}
\end{equation}
and the system (with unknowns $\Delta\tilde{\boldsymbol{\varphi}}$)
\begin{equation}
	\begin{cases}
		\textbf{M}\Delta\tilde{\boldsymbol{\varphi}} = \nabla\textbf{H}^{-k}( \boldsymbol{\varphi})^\intercal \Delta\tilde{\boldsymbol{\gamma}},\\
		\textbf{H}^{-k}( \boldsymbol{\varphi}) + \nabla\textbf{H}^{-k}( \boldsymbol{\varphi})\Delta \tilde{\boldsymbol{\varphi}} = 0,
	\end{cases}
\end{equation}
where we have removed the constraint $H_{k}( \boldsymbol{\varphi}) + \nabla H_{k}( \boldsymbol{\varphi})\Delta \boldsymbol{\varphi} = 0$ from the first system; then it holds that
\begin{equation}\label{mult_proof}
	\Delta \gamma_k\cdot\left(H_{k}( \boldsymbol{\varphi}) + \nabla H_{k}( \boldsymbol{\varphi})\Delta \tilde{\boldsymbol{\varphi}}\right) \leq 0,
\end{equation}
where $\Delta \gamma_k$ are the Lagrange multipliers of the removed constraint $k$.
\end{proposition}
\begin{proof}
Subtracting the first two equations of the systems, we get:
\begin{equation*}
	\textbf{M}(\Delta\tilde{\boldsymbol{\varphi}} - \Delta\boldsymbol{\varphi}) = \nabla\textbf{H}^{-k}( \boldsymbol{\varphi})^\intercal( \Delta\tilde{\boldsymbol{\gamma}} - \Delta\boldsymbol{\gamma}^{-k}) - \nabla H_{k}( \boldsymbol{\varphi})^\intercal\Delta\gamma_k.
\end{equation*}
Then, multiplying both sides by $(\Delta\tilde{\boldsymbol{\varphi}} - \Delta\boldsymbol{\varphi})^\intercal$, we deduce that
\begin{equation*}
	0\leq(\Delta\tilde{\boldsymbol{\varphi}} - \Delta\boldsymbol{\varphi})^\intercal\cdot\textbf{M}\cdot(\Delta\tilde{\boldsymbol{\varphi}} - \Delta\boldsymbol{\varphi}) = 0  - (\Delta\tilde{\boldsymbol{\varphi}} - \Delta\boldsymbol{\varphi})^\intercal\cdot\nabla H_{k}( \boldsymbol{\varphi})^\intercal\Delta\gamma_k,
\end{equation*}
since $(\Delta\tilde{\boldsymbol{\varphi}} - \Delta\boldsymbol{\varphi})^\intercal\cdot\nabla\textbf{H}^{-k}( \boldsymbol{\varphi})^\intercal = \textbf{H}^{-k}( \boldsymbol{\varphi})^\intercal - \textbf{H}^{-k}( \boldsymbol{\varphi})^\intercal = 0.$ Finally, using that $\nabla H_{k}( \boldsymbol{\varphi})\Delta \boldsymbol{\varphi} = -H_{k}( \boldsymbol{\varphi})$, and rearranging terms we get
\begin{equation*}
	0 \leq -\Delta\tilde{\boldsymbol{\varphi}}^\intercal \nabla H_{k}( \boldsymbol{\varphi})^\intercal\Delta \gamma_k - H_{k}( \boldsymbol{\varphi})\Delta \gamma_k.
\end{equation*}
From here (\ref{mult_proof}) follows easily.
\end{proof}

\begin{corollary}
If a constraint in the working set has a negative Lagrange multiplier, when it is taken out of the system and put in the observation set, it becomes feasible. Conversely, when a constraint in the observation set is infeasible and we sent it to the active set, its associated Lagrange multiplier is positive. 
\end{corollary}

\remark The heuristic that is usually followed to decide which constraint to remove or to add is: delete from the working set the constraint with the most negative Lagrange multiplier and add to the working set the constraint from the observational set that is being most violated (the most negative one). 

\medskip

To finish this section we study the case of linearly dependent constraints. This is relevant since in general, we do not want to introduce linearly dependent constraints into the system because they give raise to (near) singular matrices.

\begin{lemma}\label{indep_restris}
If a constraint $G$ in the observation set can be written as a linear combination of constraints of the working set, i.e. $G(\boldsymbol{\varphi}) = \sum\alpha_i H_i(\boldsymbol{\varphi})$, then the linearized constraint is feasible $G(\boldsymbol{\varphi}) + \nabla G(\boldsymbol{\varphi})\Delta\boldsymbol{\varphi}  \geq 0$.
\end{lemma}

\begin{proof}
\begin{equation*}
	G(\boldsymbol{\varphi}) + \nabla G(\boldsymbol{\varphi})\Delta\boldsymbol{\varphi} = G(\boldsymbol{\varphi}) + \sum\alpha_i \nabla H_i(\boldsymbol{\varphi})\Delta\boldsymbol{\varphi} = G(\boldsymbol{\varphi}) - \sum\alpha_i H_i(\boldsymbol{\varphi}) = 0.
\end{equation*}
\end{proof}
\remark \label{indep_apanyo} The previous lemma ensures that in general, we do not send linearly dependent constraints from the observation set to the working set. Nevertheless, it is possible to have a degenerate case where the constraints are not linearly dependent but their gradients are. In symbols, this would mean that a constraint in the observation set satisfies $G(\boldsymbol{\varphi}) + \nabla G(\boldsymbol{\varphi})\Delta\boldsymbol{\varphi}  \leq 0$ and moreover $\nabla G(\boldsymbol{\varphi}) = \sum\alpha_i \nabla H_i(\boldsymbol{\varphi})$. What we do then is to introduce $G$ in the working set while removing the $H_i$ with the greatest $\alpha_i\neq 0$ in absolute value. This new working set is linearly independent (otherwise it would contradict the assumption that the original working set without $G$ was linearly independent) and the process can continue.

\subsection{Factorization of the matrix system}

Every time that a constraint goes from the working set to the observation set (or vice versa), i.e. when the index sets $\mathcal{W}$ and $\mathcal{O}$ are updated, the Lagrange multipliers must be recomputed, i.e. a linear system must be solved in order to find the solution of (\ref{sist_work}). 

\begin{equation}\label{sist_work}
\begin{cases}
	\textbf{M}\cdot\Delta \boldsymbol{ \boldsymbol{\varphi}}_{j+1} = -\nabla\textbf{C}( \boldsymbol{\varphi}_{j})^\intercal\Delta \boldsymbol{\lambda}_{j+1} + \nabla\textbf{H}( \boldsymbol{\varphi}_{j})^\intercal\Delta \boldsymbol{\gamma}_{j+1} -\mu\textbf{V}(\dot{ \boldsymbol{\varphi}}_j)^\intercal\Delta\boldsymbol{\beta}_j,\\
	\textbf{C}( \boldsymbol{\varphi}_{j}) + \nabla\textbf{C}( \boldsymbol{\varphi}_{j})\Delta \boldsymbol{\varphi}_{j+1} = 0,\\
	H_i( \boldsymbol{\varphi}_{j}) + \nabla H_i( \boldsymbol{\varphi}_{j})\Delta \boldsymbol{\varphi}_{j+1} = 0 \text{ for } i\in\mathcal{W}.
\end{cases}
\end{equation}

In order to ease readability we will include inextensibility constraints and the contact constraints of the working set in only one function denoted by $\textbf{G}^\intercal = [\textbf{C}^\intercal,\textbf{H}^\intercal]$. Now, since in general only one constraint will be entering or exiting at the time, the linear systems that we have to solve are almost identical with the exception of a few rows and columns. That is why, the use of factorizations becomes an important tool to achieve efficiency. The linear system we need to solve to find the multipliers is:

\begin{equation*}
\left(\nabla\textbf{G}( \boldsymbol{\varphi}_j) \textbf{M}^{-1}\nabla\textbf{G}(\boldsymbol{\varphi}_j)^\intercal\right)\Delta\boldsymbol{\zeta}_{j+1} = -\textbf{G}(\boldsymbol{\varphi}_j) - \nabla\textbf{G}(\boldsymbol{\varphi}_j)\textbf{M}^{-1}\textbf{f}_{\mu}(\boldsymbol{\varphi}_j),
\end{equation*}
where $\Delta\boldsymbol{\zeta}_{j+1}^\intercal = [\Delta\boldsymbol{\lambda}_{j+1}^\intercal,\Delta\boldsymbol{\gamma}_{j+1}^\intercal]$. Hence the system matrix is positive definite (since $\textbf{M}$ is positive definite because it is the mass matrix); therefore we can use Cholesky decomposition \cite{Golub1983MatrixC}, provided our constraints are linearly independent (see again Lemma \ref{indep_restris} and Remark \ref{indep_apanyo}). Every time a constraint enters or exits the working set, the Cholesky decomposition of the system matrix can be efficiently updated without recomputing the factorization from scratch (see, e.g. \cite{Seeger2004LowRU,Davis2009DynamicSI})

\subsection{Detailed algorithm for collisions}

To finish this section we give a detailed description of the full numerical algorithm written in pseudo-code in Algorithm \ref{algo_coli}. 

\begin{algorithm}[htb!]
\begin{algorithmic}[1]
	\Require{$\boldsymbol{\varphi}^n,\dot{\boldsymbol{\varphi}}^n$}
	\State {$\boldsymbol{\varphi}_0 \gets \text{unconstrained}(\boldsymbol{\varphi}^n,\dot{\boldsymbol{\varphi}}^n,\dots),\;j\gets0$}
	\Comment(\textit{Implicit Euler integration step})
	\State{ $\mathcal{C}_0 = \text{Collisions}_{\omega}\left(\boldsymbol{\varphi}^{n} \rightarrow_{dt}\boldsymbol{\varphi}_{0}\right)$}
	\Comment(\textit{Self-collision constraints})
	\State {$\mathcal{W} \gets \{i: G_i(\boldsymbol{\varphi}^n) = 0\},\; \mathcal{O} \gets \mathcal{W}^c$}
	\Comment(\textit{Working and observation sets})
	\State {$\boldsymbol{J} \gets \text{gradient}(\boldsymbol{\varphi}_0,\mathcal{C}_0,\mathcal{W},\dots)$}
	\Comment(i.e. $\nabla G_i:\; i\in \mathcal{W}$)
	\State {$\boldsymbol{L} \gets \text{cholesky}\left(\textbf{J}\cdot\textbf{M}^{-1}\cdot\textbf{J}^\intercal\right)$}
	\While{$\max|\textbf{C}(\boldsymbol{\varphi}_j)| \geq \epsilon_0$ or $\min\textbf{H}(\boldsymbol{\varphi}_j) \leq -\epsilon_1$} 
	\State {$[\Delta\boldsymbol{\lambda},\Delta\boldsymbol{\gamma}] = \text{multipliers}(\boldsymbol{\varphi}_j,\textbf{L})$}
	\If {$\min(\Delta\boldsymbol{\gamma})\geq 0$}
	\State{$\Delta\boldsymbol{\varphi}_{j+1} \gets \text{increment}(\Delta\boldsymbol{\lambda},\Delta\boldsymbol{\gamma},\textbf{J})$}
	\If {$H_i( \boldsymbol{\varphi}_{j}) + \nabla H_i( \boldsymbol{\varphi}_{j})\Delta \boldsymbol{\varphi}_{j+1} \geq 0 \text{ for } i\in\mathcal{O}$}
	\State{$ \boldsymbol{\varphi}_{j+1} \gets \boldsymbol{\varphi}_j + \Delta \boldsymbol{\varphi}_{j+1}$}
	\State{ $\mathcal{C}_{j+1} = \text{Collisions}_{\omega}\left(\boldsymbol{\varphi}^{n} \rightarrow_{dt}\boldsymbol{\varphi}_{j+1}\right),\; \mathcal{O} \gets \mathcal{O}\cup \mathcal{C}_{j+1}$}
	\State {$\boldsymbol{J} \gets \text{gradient}(\boldsymbol{\varphi}_{j+1},\cup_k\mathcal{C}_k,\mathcal{W},\dots)$}
	\State {$\boldsymbol{L} \gets \text{cholesky}\left(\textbf{J}\cdot\textbf{M}^{-1}\cdot\textbf{J}^\intercal\right),\; j\gets j+1$}
	\Else
	\State{$i_{in} \gets \text{indmin}_{i\in\mathcal{O}}(H_i( \boldsymbol{\varphi}_{j}) + \nabla H_i( \boldsymbol{\varphi}_{j})\Delta \boldsymbol{\varphi}_{j+1})$}   
	\State{$\mathcal{O} \gets \mathcal{O}\setminus i_{in},\; \mathcal{W} \gets \mathcal{W}\cup i_{in}$ }   
	\State{$\textbf{L} \gets \text{update}(\textbf{L},i_{in})$}   
	\EndIf
	\Else
	\State{$i_{out} \gets \text{indmin}(\Delta\boldsymbol{\gamma})$}   
	\State{$\mathcal{O} \gets \mathcal{O}\cup i_{out},\; \mathcal{W} \gets \mathcal{W}\setminus i_{out}$ }   
	\State{$\textbf{L} \gets \text{update}(\textbf{L},i_{out})$}   
	\EndIf
	\EndWhile 
	\State{$\boldsymbol{\varphi}^{n+1}\gets\boldsymbol{\varphi}_{j^*},\;\dot{\boldsymbol{\varphi}}^{n+1}\gets\frac{\boldsymbol{\varphi}^{n+1}-\boldsymbol{\varphi}^{n}}{dt}$}\\		
	\Return{$\boldsymbol{\varphi}^{n+1},\dot{\boldsymbol{\varphi}}^{n+1}$}
	\caption{Collisions active-set algorithm}\label{algo_coli}
\end{algorithmic}
\end{algorithm}

\remark We now make some comments about Algorithm \ref{algo_coli}:

\begin{enumerate}
\item[1] The working set is always initialized at least with the inextensibility constraints, but we can also add the active contact constraints from the previous time-step $t_{n-1}$.
\item[2] We have not explicitly written the friction force, but it obviously comes up in the computation of the multipliers and the increment (lines 7 and 9, see Equations (\ref{sist_work})).
\item[3] Note that after successfully finding the active set (line 10), for the next step $j+1$, we do not change the working set $\mathcal{W}$. Only the observation set $\mathcal{O}$ is updated with the possible new self-collision constraints found (line 12).
\item[4] When there is a negative multiplier (line 20), note that we take out from the system the constraint with the smallest (most negative) multiplier. Likewise, when one of the constraints in the observation set must be introduced (line 15), we choose the one that is being most violated.
\end{enumerate}

\subsection{Similarities and differences with common active-set methods}

Now that we have presented the full algorithm we use to solve the quadratic problems, we can talk in more detail about how it compares to standard active-set methods like the one described in \cite{Nocedal1999NumericalO}. The main difference was already mentioned: to our knowledge, all active-set methods require that one begins at a feasible point and then iterates from there. This has the disadvantage that one must find a feasible point to begin with, e.g. solving a linear program with equality and inequality constraints. On the other hand, those classic methods have the advantage that all the constraints in the observation set are kept non-violated, and this allows one to take smaller steps towards the solution when all the multipliers are positive (potentially causing the algorithm to converge faster). Since we are solving so many relatively large sparse quadratic problems in a row, we have found that the requirement of starting at a feasible point is way more expensive than employing the novel algorithm here presented. This is the case because lower-rank updates of sparse Cholesky decompositions can be carried out very efficiently.  

\section{Qualitative evaluation of the collision model}\label{sec_eval_cols}
In this section, we present several experiments to test our collision model. They are qualitative in nature, i.e. we show that our simulator is capable of dealing with such scenarios. We show that our modelization of friction is effective in static (cylinder experiment, Section \ref{sec_cilindro}) and dynamic (rotating sphere experiment, Section \ref{sec_esfera_friccion}) settings, that we can easily include collisions with sharp objects (Section \ref{sec_cuspides}) and that we can simulate complicated folding sequences of cloth with non-trivial topologies (shorts experiment, Section \ref{sec_pants}). The second and the third experiments are challenging scenarios suggested by \cite{Li2021CodimensionalIP} as challenging tests for a robust cloth collision model.  

\subsection{Frictional cylinder}\label{sec_cilindro}

\begin{figure}[htb!]
	\centering
	\includegraphics[width=1\linewidth]{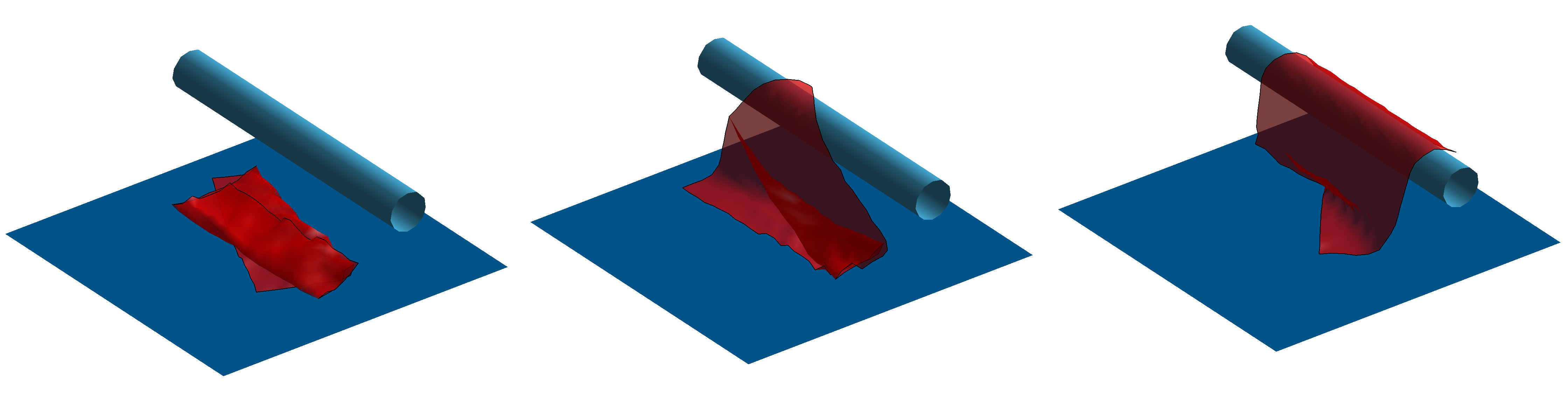}
	\caption{\label{cilindro}Final frame ($t=2$ seconds) of 3 separate simulations of the fall of a sheet of cloth on top of an off-center cylinder. All the physical parameters are kept constant but friction, which varies among $\mu\in\{0.2,0.4,0.55\}$.}
\end{figure}

This is the most basic of the four experiments: a flat sheet of cloth falls on top of a frictional cylinder during 2 seconds. In Figure \ref{cilindro} we show the final configuration of the textile for $t = 2$. All the physical parameters are kept constant but friction, which varies among $\mu\in\{0.2,0.4,0.55\}$. The cylinder is 30 cm off-center (with respect to the center of mass of the cloth whose measures are $130 \text{ cm} \times 130 \text{ cm}$) and the textile is slightly rotated (20 degrees with respect to the $z$-axis). This means that in the absence of friction (or with a small friction coefficient), the cloth collides with the cylinder and then falls to the floor. We show with this scenario that our implementation of friction is effective and can handle scenarios with persistent contact. 
In the first image of Figure \ref{cilindro}, we have that $\mu = 0.2$ is too small and therefore the cloth falls onto the floor. In the second image, the friction $\mu = 0.4$ is somehow bigger and the cloth can be seen still in the process of falling but at a later stage, which shows that the friction forces have acted and delayed the fall. Finally, in the third image with $\mu = 0.55$, the friction is high enough so that the sheet lies stably on top of the cylinder. For a video of the three simulations, see \url{https://youtu.be/_nh-ejHcJAg}.

\subsection{Rotating sphere}\label{sec_esfera_friccion}

\begin{figure}[htb!]
	\centering
	\includegraphics[scale=0.75]{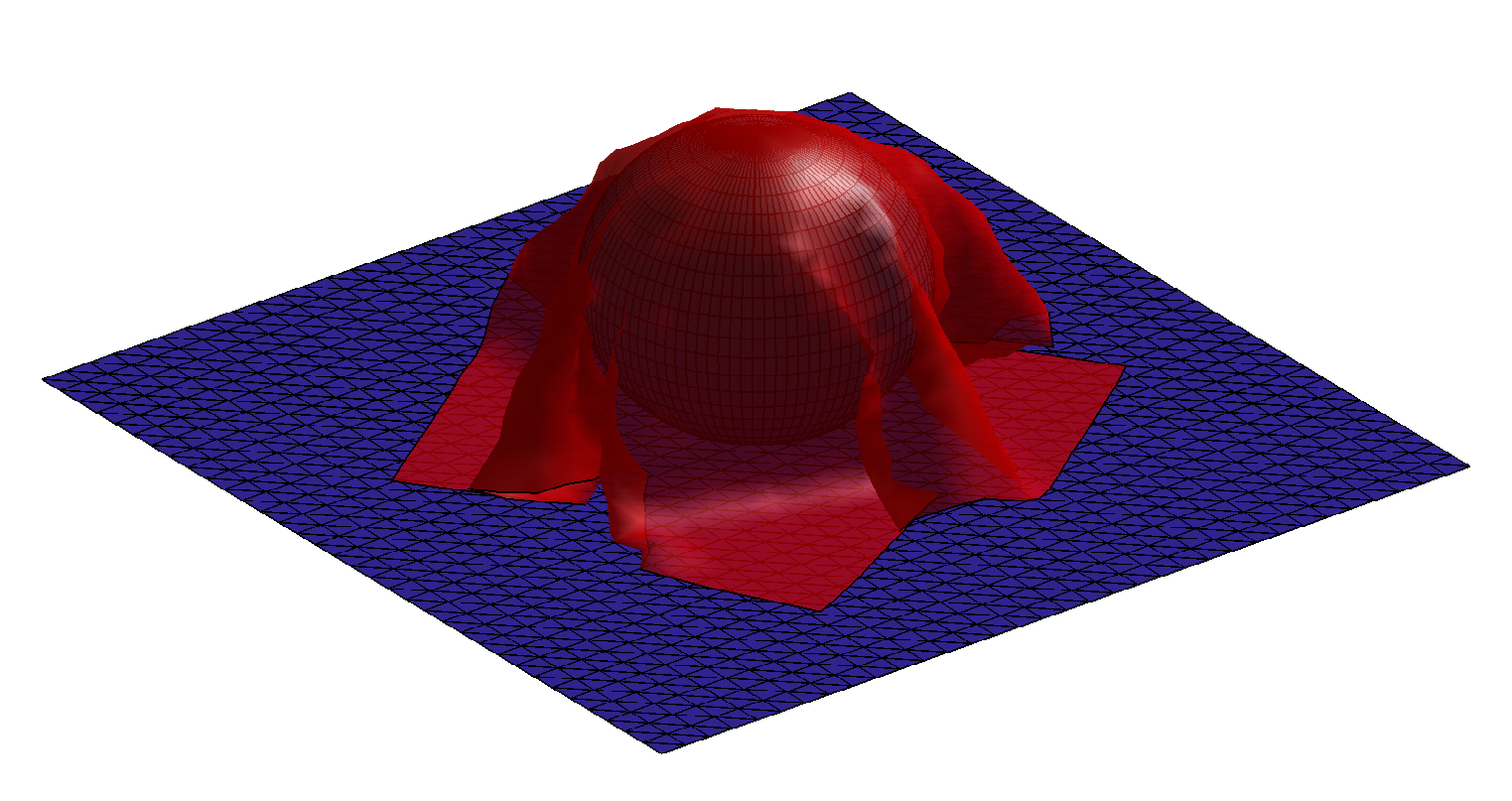}
	\caption{\label{esferas}Final frame ($t=2.5$ seconds) of the collision of a sheet of cloth with a frictional sphere and the floor. One second after the textile has fallen, the sphere performs half a rotation along the $z$-axis during one second. }
\end{figure}

In this second experiment, we simulate the collision of a sheet of cloth with a frictional sphere and the floor. The cloth measures $190 \text{ cm} \times 190 \text{ cm}$ whereas the sphere has a radius of 35 cm. One second after the textile has fallen, the sphere performs half a rotation along the $z$-axis during one second. The discrepancy in size is intentional so that after the fall the textile is also in contact with the floor and can then wrap around the sphere. In Figure \ref{esferas} we can see the final frame of the simulation at $t = 2.5$ seconds. For a full video of the simulation, see \url{https://youtu.be/C8izvprEcKk}.

\smallskip

In order to simulate properly the dynamical friction between the cloth and the sphere, we must take into account the speed of the rotation for the points of the mesh that are in contact with the sphere. In practice this means that when computing at every iteration $j$ of our solver the tangent velocities, we must account for a new term:     

\begin{equation*}
	\textbf{V}(\dot{\boldsymbol{\varphi}_j}) = \dot{\boldsymbol{\varphi}_j} - \langle\dot{\boldsymbol{\varphi}_j},\textbf{n}\rangle\cdot \textbf{n} - \textbf{v}^{sphere},
\end{equation*}
where $\textbf{n}$ is the outwards normal to the sphere and $\textbf{v}^{sphere}$ is its speed at the current time step (as before the tangent velocities are afterwards normalized). 

\medskip

In order to obtain an interesting behavior of the simulation, it is important to calibrate carefully the interplay between the friction with the floor and with the sphere. We select $\mu_{sphere} = 0.5$ and  $\mu_{floor} = 0.4$, so that the cloth follows the rotation of the sphere but with considerable resistance from the floor. 


\subsection{Collision with a sharp obstacle}\label{sec_cuspides}
\begin{figure}[htb!]
	\centering
	\includegraphics[width=0.75\linewidth]{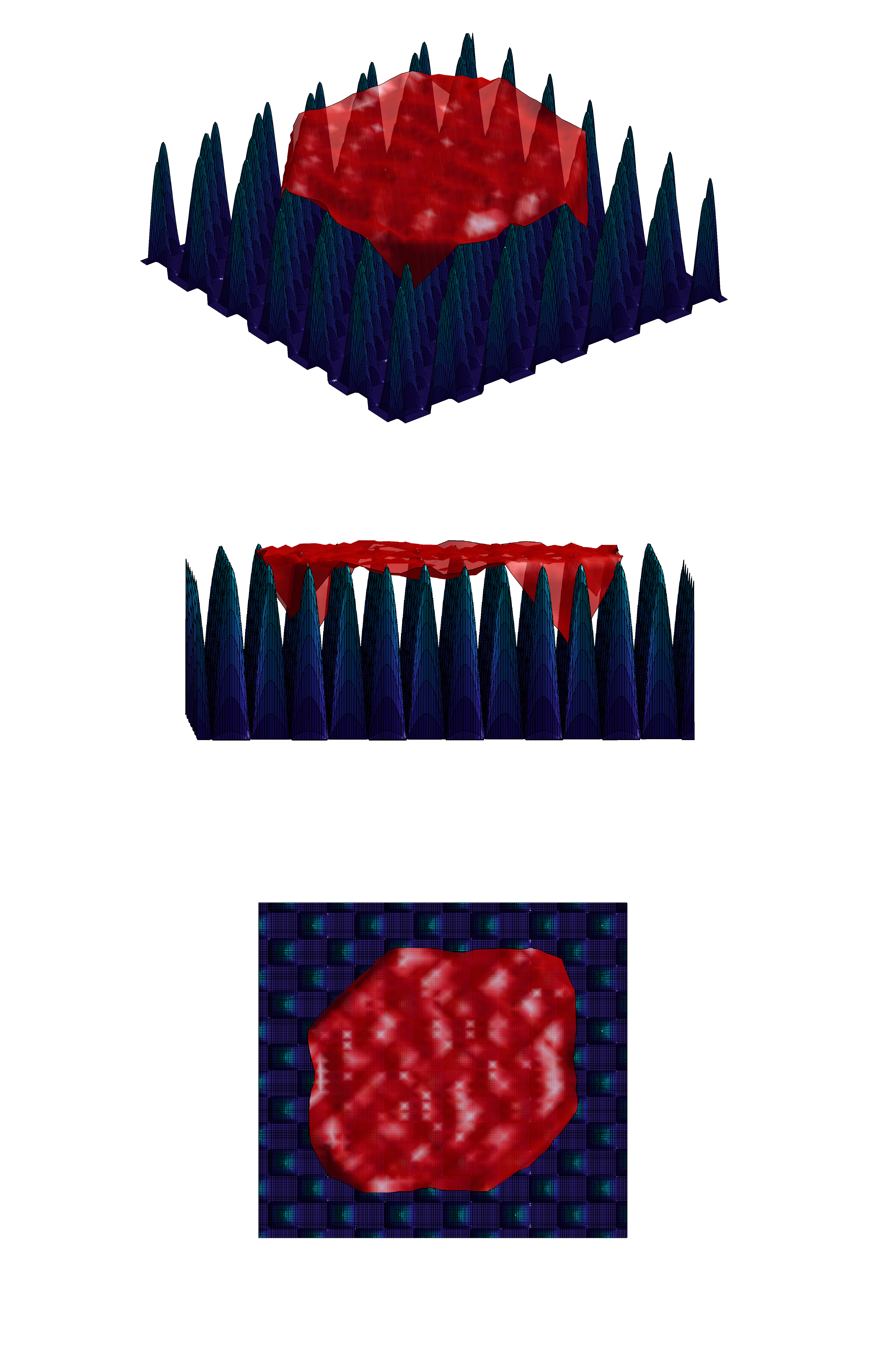}
	\caption{\label{cuspides}Simulated final frame of cloth's collision with a collection of needle-like obstacles seen from 3 different angles. The cusps must be taken into account separately from the rest of the surface and are treated with the same algorithm we treat self-collisions.}
\end{figure}

In this third experiment, we simulate the collision of a piece of cloth with a collection of needle-like obstacles. They are given by the set of implicit equations:

\begin{equation}\label{huevera}
	\textbf{H}(\boldsymbol{\varphi}) = c_1c_2\textbf{z} - \sin(c_1\textbf{x})\sin(c_1\textbf{y}),
\end{equation} 
where we take $c_1 = 20$ and $c_2 = 0.075$ (see Figure \ref{cuspides}).

The interest of this scenario lies in the fact that it is not enough to impose the previous equation (\ref{huevera}) as a hard constraint (like we did with the sphere and the cylinder); but that we need in addition to take into account the cusps of the surface. These difficulties are typical for most physical simulators and they arise when the obstacles we are simulating present characteristics of lower dimensional objects (e.g. a really thin cylinder or the cusps in this case). It is easy to see that the cusps are given by: 
\begin{equation}
	x = \frac{2\pi m \pm \frac{\pi}{2}}{c_1}, \quad y = \frac{2\pi m \pm \frac{\pi}{2}}{c_1}, \quad z = \frac{1}{c_1c_2},
\end{equation} 
where $m\in\mathbb{Z}$.

\medskip

Let us denote them by $\{q_1,\dots q_f\}$. Then, for every iteration $j$ of the solver, similarly like we do with self-collisions, we must check if a collision occurred during the motion $\boldsymbol{\varphi}^{n}\rightarrow_{dt}\boldsymbol{\varphi}_{j}$ between these cusps and the (triangular) faces of our meshed cloth. This means that for every detected collision, in the next iteration $j+1$ we must add a constraint of the form:  

\begin{equation*}
	\langle q_i - \pi(x_1,x_2,x_3) ,\nu\rangle \geq 0,
\end{equation*}
where $q_i$ is the corresponding cusp, $x_i$ are the 3 corners of the triangle, $\pi(x_2,x_3,x_4) = ux_1 + vx_2 + wx_3$ is the closest point between the face and $q_i$, and $\nu$ is the normal vector to the triangle. 

\medskip

The values $\nu,u,v,w$ are (like in the case for self-collisions) constant in time, and are computed with the positions given by $\boldsymbol{\varphi}_{j}$. The normal vector $\nu$ is oriented such that $H_i(\boldsymbol{\varphi}^{n})\geq 0$.

\remark As with self-collisions we consider cloth's thickness in practice by imposing $H_k(\boldsymbol{\varphi})\geq\tau_0 > 0$. Moreover, as before this thickness is taken into account in the detection process (see Section \ref{grosor}). 

\medskip

In Figure \ref{cuspides} we can observe the result of the simulation from three different viewpoints. The cloth lies stably on top of the cusps without any noticeable artifacts. For a video of the simulation, see \url{https://youtu.be/z7l_O_nSfrM}.

\subsection{Folding sequence of short pants}\label{sec_pants}
In this final experiment, we simulate the dynamical folding of a pair of shorts on top of a table. In order to do so, we control two nodes at the top of the shorts. 
The first part of the motion is performed fast enough so that the shorts have sufficient momentum to lay partially flat on top of the table after lowering them. Finally, the fold is completed by dropping the top two corners on top of the leg loops. 


\begin{figure}[H]
	\centering
	\includegraphics[scale=0.24]{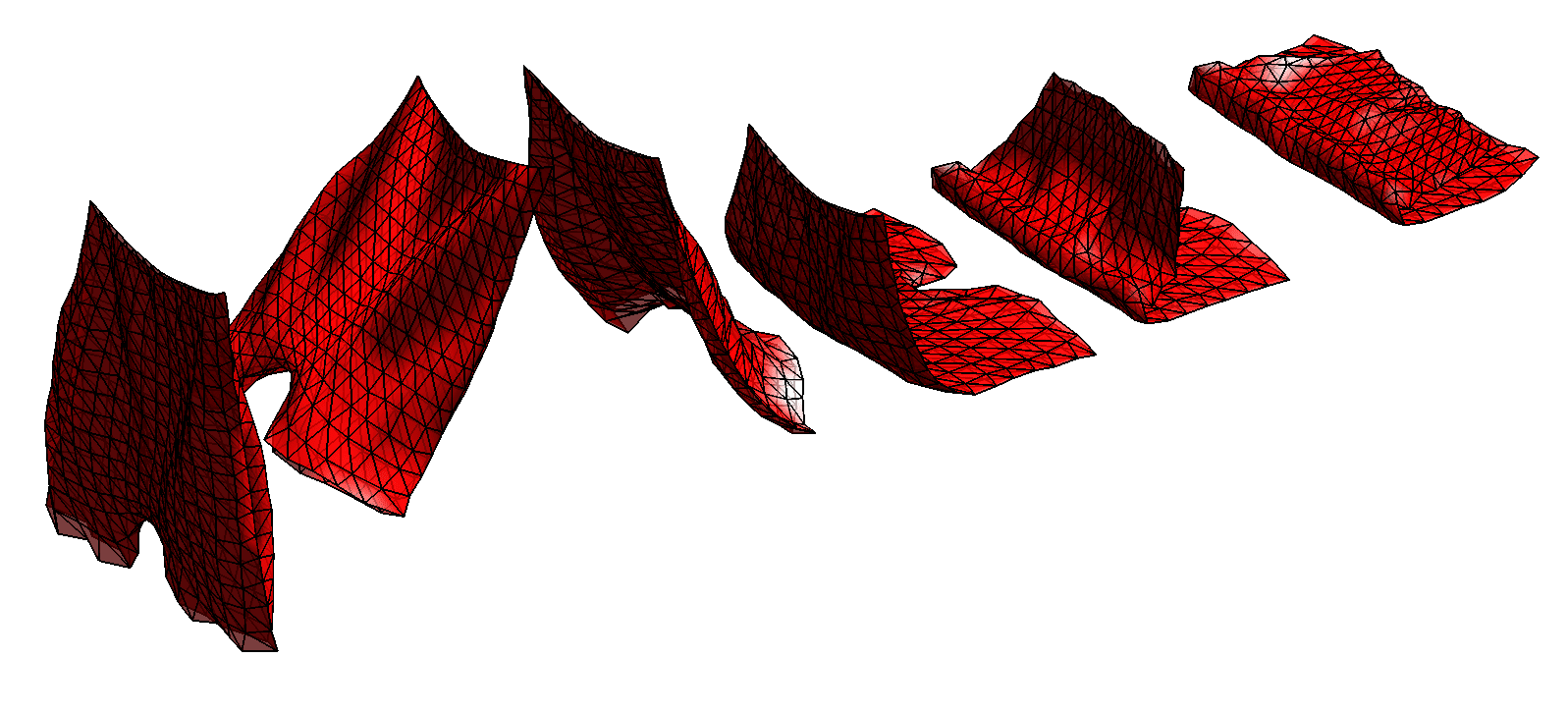}
	\caption{\label{shorts_folded}Simulated sequence of the dynamical folding of a pair of shorts. The first part of the motion (frames two and three) is performed fast enough so that the shorts lay partially flat on top of the table after a lowering phase (frame four). The final fold is completed by dropping the top two corners on top of the leg loops (frames five and six).}
\end{figure}

In Figure \ref{shorts_folded} we depict six frames of the simulation. Notice how crucial is the well-functioning of the self-collisions algorithm for a realist outlook of the whole folding sequence. For a full video of the simulation, see \url{https://youtu.be/2gdnjUICb0g}.

\section{Experimental validation of the collision model} \label{sec_expvalidation}

The definitive test for a model of cloth is its comparison to reality. Textile engineers have focused on such tasks, to the point of developing specialized testing equipment. But the goal of their study has always been local properties of cloth, such as elasticity parameters, which are tested in static scenarios (e.g. \cite{Hu04,Wang:2011:DDE,Miguel2012,Clyde2017}). Other, more recent lines of research such as \cite{Rasheed2020LearningTM,Rasheed2021AVA} focus on estimating friction coefficients using non-intrusive video images. To the knowledge of the authors, none of the models previously mentioned has been able to compare its results with dynamic motions of textiles involving collisions. 

\medskip

To perform this set of experiments, we use a \textit{Motion Capture System}, and record two collision scenarios of four textiles. In one of the scenarios, the fabrics are laid dynamically on top of a table in a putting-a-tablecloth fashion. In the other, they are hit by a cylindrical stick four times at various places and with different strengths. The goal is to assess the accuracy of the collision and friction model previously developed. For the first scenario, we find the optimal friction parameter for both a high and a low friction case and study the stability of the model with respect to this parameter. For the hitting experiment, we find again the optimal parameters of the model and also check computational times comparing our active-set solver with a standard interior-point method.

\subsection{Cloth's materials and sizes}

For the experiments in this validation, we employ four cloth materials of size DIN A2 (0.42 x 0.594 m). Before performing the experiments they were ironed as to remove all considerations of plasticity from the validation process. In Table \ref{table_telas} we can see the density and type of all the fabrics and some typical examples of garments made from them. 

\begin{table}[htb!]
	\centering
	\begin{tabularx}{0.7\textwidth}{Xccc} \toprule
		\tableheadline{Fabric} & \tableheadline{Density ($\text{kg}\cdot \text{m}^{-2}$)} & \tableheadline{Size} & \tableheadline{Examples} \\ \midrule
		Polyester & 0.1042 & A2 & Silk-like.\\
		Wool & 0.1804 & A2 & Formal suit. \\ 
		Denim & 0.3046 & A2 & Jeans. \\ 
		Stiff-cotton & 0.3046 & A2 & Sack. \\ 
		\bottomrule
	\end{tabularx}
	\caption{Density, sizes and examples of all the materials used in all the experiments.}
	\label{table_telas}
\end{table}

\subsection{Recording setting}

\begin{figure}[htb!]
	\centering
	\includegraphics[width=0.7\linewidth]{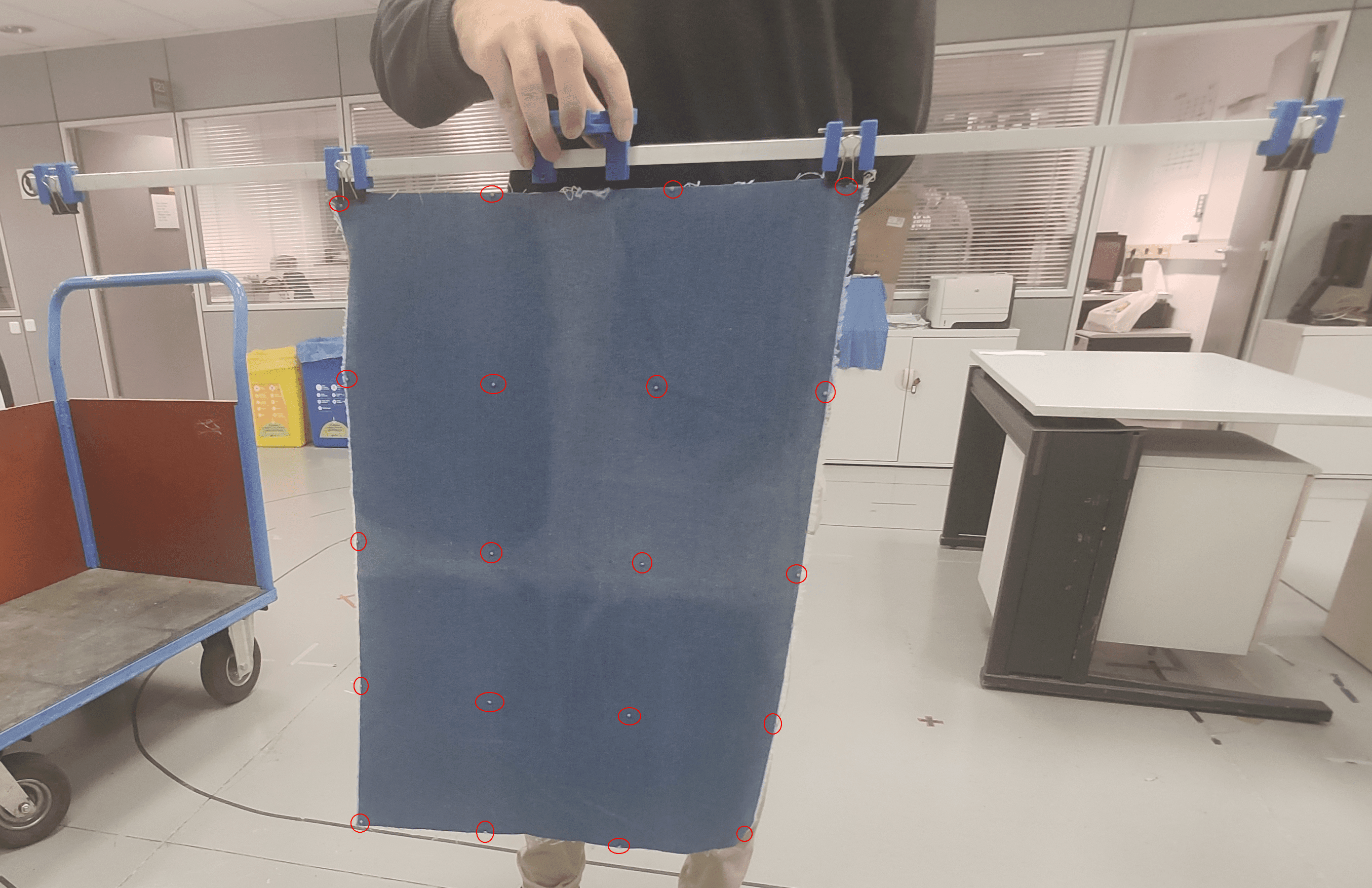}
	\caption{\label{markers} Reflective markers attached to the denim sample (encircled in red). The markers are very small, with a diameter of $3$ mm and a weight of $0.013$ g. We use 20 reflective markers.}
\end{figure}

\begin{figure}[htb!]
	\centering
	\includegraphics[width=0.95\linewidth]{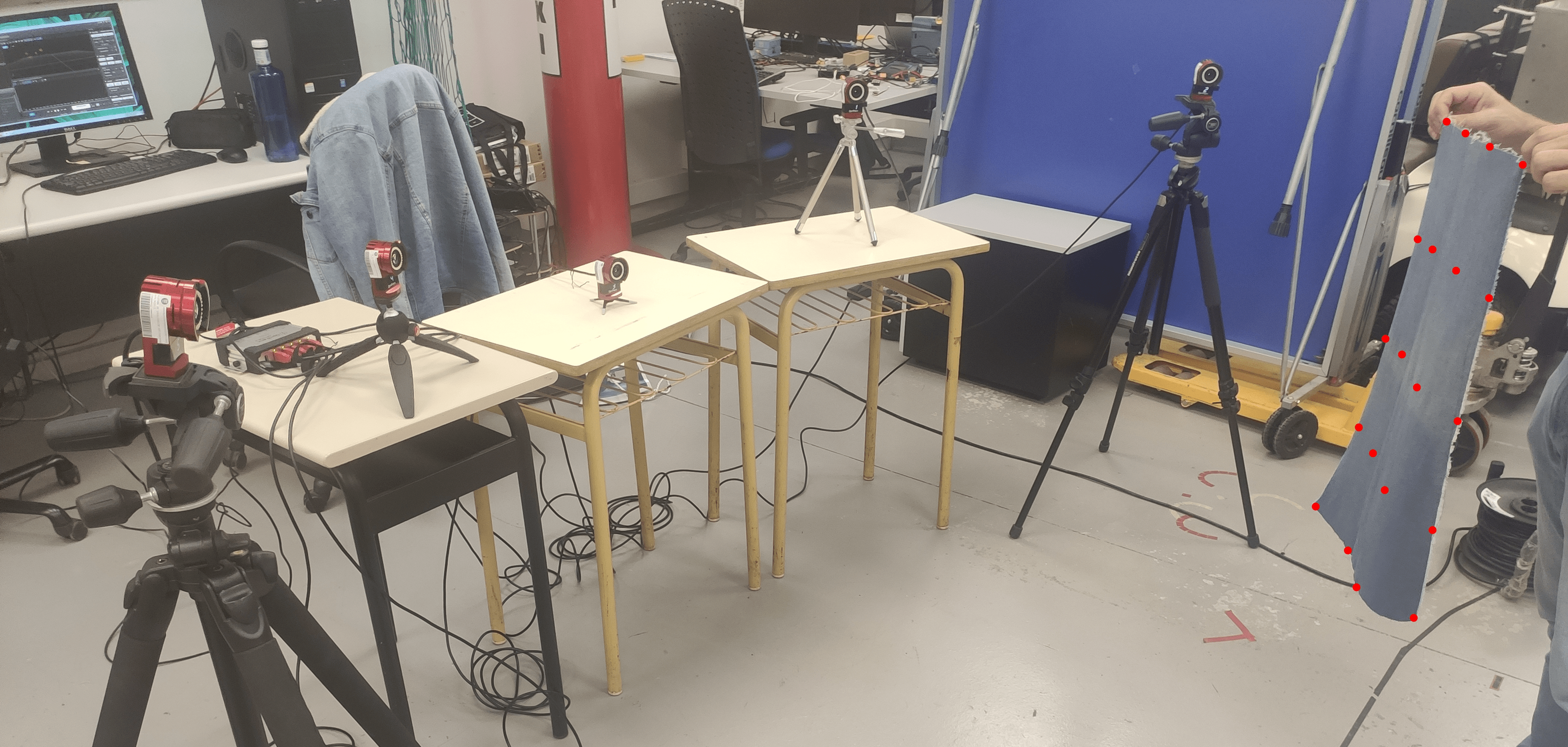}
	\caption{\label{setup_opti} Setup used to record the motion of the textiles: 5 cameras surround the scene so that every marker (highlighted in red in the photo) is visible to at least 2 cameras at the same time. This ensures that the system can be certain of the 3D position of the marker.}
\end{figure}

To record the motion of the textiles a system of cameras detects and tracks reflective markers that are hooked on the cloth (see Figure \ref{markers}). These markers, with a diameter of $3$ mm and a weight of $0.013$ g, reflect infrared light, so the cameras are able to follow their motion through space. We use hardware and software from the manufacturer \textit{NaturalPoint Inc}: five \textit{Optitrack Flex 13} cameras surround the scene we wish to record (see Figure \ref{setup_opti}) and afterwards the recordings are processed with the software \textit{Motive}. This combination of software and hardware offers sub-millimeter marker precision, in most applications less than $0.10$ mm according to the manufacturers. 

\smallskip

This technology has been extensively used to track the motion of rigid and articulated bodies (e.g. human movements by following the trajectories of all joints). Nevertheless, its use for deformable objects has been less common since the weight of the markers could affect the dynamics of the object. This does not happen in our case since the markers we use are so light and small that account for less than $1\%$ of cloth's weight even for the lightest materials.

\medskip

We use 20 reflective markers, which are placed equidistantly in order to obtain a faithful representation of the dynamics of the fabrics. In contrast to the experiments done in \cite{COLTRARO2022}, this time the motions are performed by a human. This introduces more uncertainty, since every movement has its own unique variabilities. 

\remark Since we have 5 cameras surrounding a scene, we can record more varied and faster movements without losing track of the textiles, as opposed to, e.g. using just one depth camera. Nevertheless, some markers are lost some of the time (especially with fast or abrupt movements), for instance when the textiles deform so much that the corners are no longer visible to the cameras. We have taken care that in our recordings these disappearances only happen for short periods of time. 

\subsection{Parameter fitting}\label{sec_vali_cols}

We denote the sequence of positions of the recorded fabric's nodes given by the motion capture system by $\{\boldsymbol{\phi}^0, \boldsymbol{\phi}^1,\dots, \boldsymbol{\phi}^m\}$ and as already said, the simulated sequence obtained by using Algorithm \ref{algo_coli} by $\{ \boldsymbol{\varphi}^0, \boldsymbol{\varphi}^1,\dots, \boldsymbol{\varphi}^m\}$. This sequence is obtained by taking $\boldsymbol{\varphi}^0 = \boldsymbol{\phi}^0$ and using the same recorded trajectories of the two upper corners for the simulations. In order to validate the realism of our collision model, we fit three parameters: $\alpha$ (Rayleigh damping, this damps long oscillations) and $\delta$ (virtual mass, this models aerodynamics), and  $\mu$ (friction coefficient). The first two parameters were introduced in \cite{COLTRARO2022} as an accurate way to describe the dynamics of inextensible sheets of cloth without collisions, and $\mu$ was introduced in Section \ref{sec_cols_fric}. In order to obtain their optimal value, we minimize the mean along time of the absolute error:

\begin{equation}\label{error_rel_mean}
	\sum_i e_i(\delta,\alpha,\mu) = \sum_i\sqrt{|| \boldsymbol{\varphi}^i(\delta,\alpha,\mu) -  \boldsymbol{\phi}^i||_{\textbf{M}}^2},
\end{equation}
where $||\cdot||_{\textbf{M}}$ is the $L^2$ norm with respect to the matrix $\textbf{M}$  (i.e. $||\textbf{x}||_{\textbf{M}}^2 = \textbf{x}^\intercal\cdot\textbf{M}\cdot\textbf{x}$). All other physical parameters (e.g. bending) are set to $0$ except for $\rho$ which is set to its corresponding value of Table \ref{table_telas}. For a justification of this choice of parameters, see \cite{COLTRARO2022}. For the simulations we consider a refinement of the initial mesh $4\times 5$ given by the markers, that is, we employ a $7\times 9$ resolution.

\medskip

As metrics to evaluate the fitting of the model, we use the (time dependent on i) absolute error:
\begin{equation}\label{error_rel}
	e_i(\delta,\alpha,\mu) = \sqrt{|| \boldsymbol{\varphi}^i(\delta,\alpha,\mu) -  \boldsymbol{\phi}^i||_{\textbf{M}}^2},
\end{equation}
and the following time-dependent (on $i$) spatial (on $j$) standard deviation: 

\begin{equation}\label{desv_tipica}
	s_i(\delta,\alpha,\mu) = \sqrt{\Var_{j\in\Nodes(S)}\left(|| {\varphi}_j^i(\delta,\alpha,\mu) -  {\phi}_j^i||_{\mathbb{R}^3}\right)}.
\end{equation}

\remark The errors are only computed at the recorded nodes. Furthermore, as mentioned before, some of the markers disappear for small periods of time, in those cases, they are simply excluded from the computation of the errors (no interpolation is performed).

\medskip

The experiments are performed by a human (with bare hands) and consist of two scenarios:

\subsection{Tablecloth scenario}\label{sec_tablecloth}

\begin{figure}[htb!]
	\centering
	\includegraphics[width=0.6\linewidth]{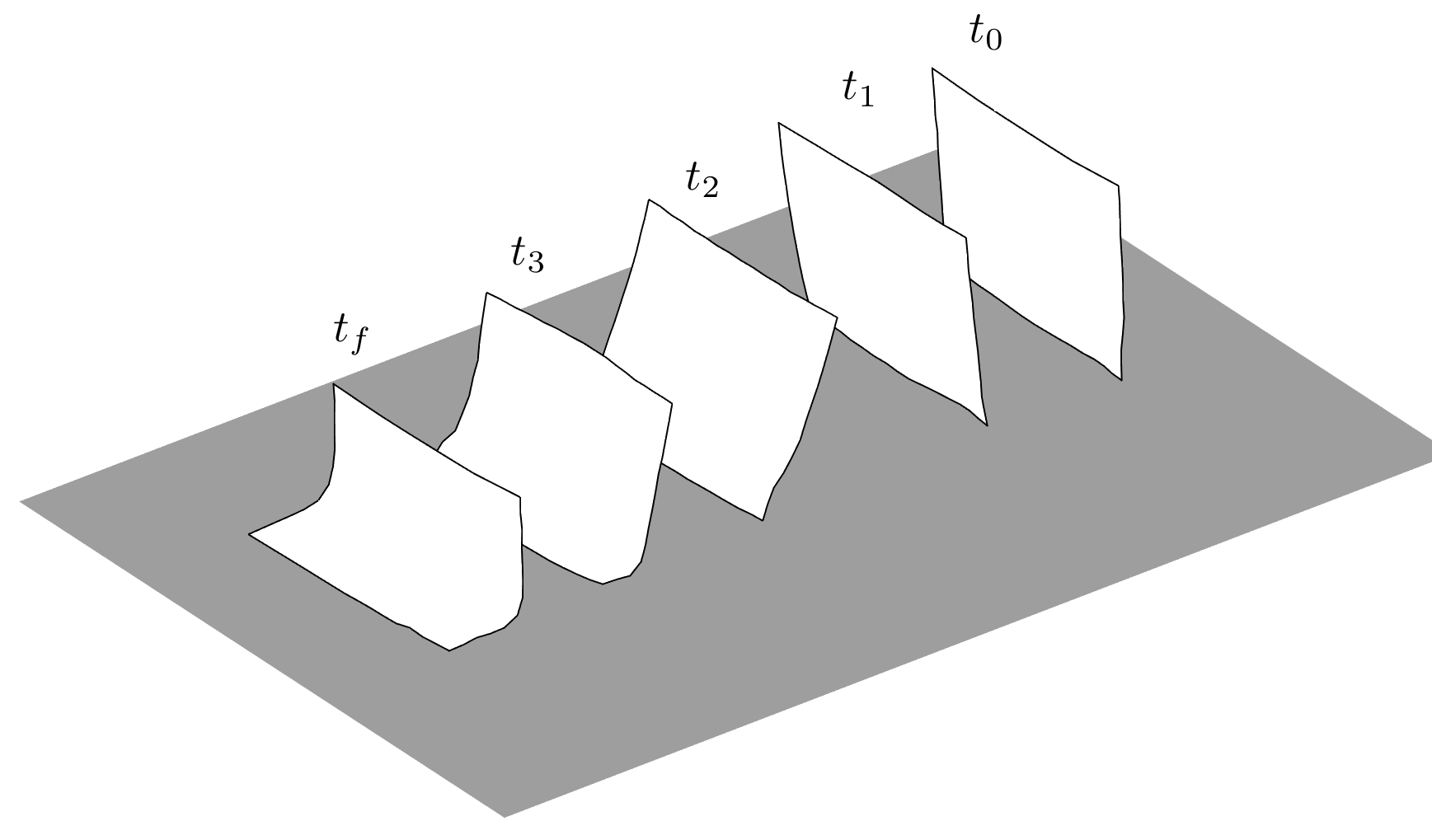}
	\caption{\label{manta_doblada}Putting a tablecloth motion sequence (right to left): the cloth starts suspended and is afterwards laid dynamically (only partially) onto the table. }
\end{figure}

The textile starts suspended at about 10 cm of height and is afterwards laid dynamically (only partially, so that half of the cloth is still suspended) onto the table (see Figure \ref{manta_doblada}). Each motion lasts approximately 4 seconds (with a frame every $dt = 0.01$ seconds) and is performed with two different surfaces as the table, one with \textit{low} friction (a raw polished table) and one with \textit{high} friction (a table with a tablecloth). The goal here is to estimate the friction coefficient $\mu$ (see Equation (\ref{KKTcoli})) for the two different surfaces and to study the sensitivity of the model with respect to friction.

\begin{figure}[htb!]
	\centering
	\includegraphics[width=0.8\linewidth]{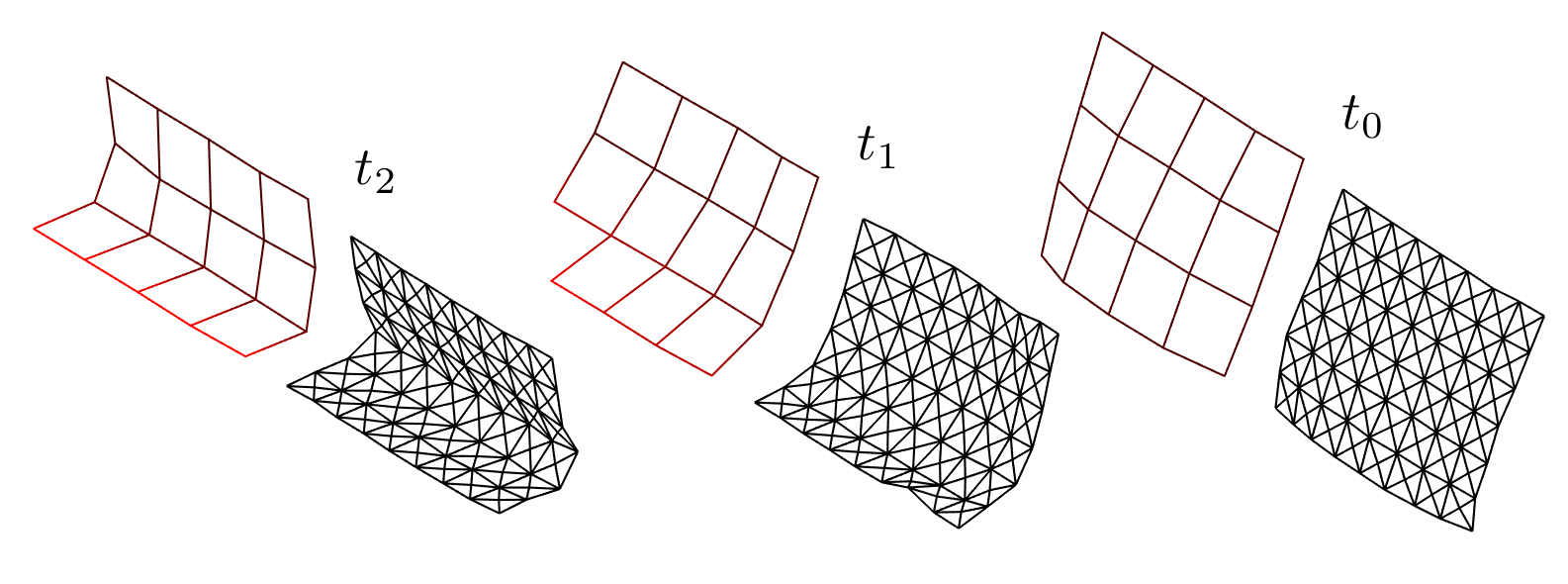}
	\caption{\label{lana_doblada}Three frames comparing the recorded tablecloth low friction scenario of A2 wool (left) with its inextensible simulation (right). The error at the three depicted frames from right to left is 0.80, 1.17, and 0.76 cm respectively; being the average error of the whole simulation 0.58 cm.}
\end{figure}

In Table \ref{tabla_fricciones} we can see the optimal values of the friction coefficients along with their optimal errors and deviations for the low and high friction scenarios. The optimal friction coefficients for the low friction case (raw polished table) were all smaller than $10^{-3}$ and that is why they were rounded up to zero on the table. For a visual comparison of the results see Figure \ref{lana_doblada} and \url{https://youtu.be/sWJcxfTwKHE}.

\begin{table}[htb!]
	\centering
	\begin{tabularx}{0.75\textwidth}{lcccccc} \toprule
		\tableheadline{Material} & 
		\tableheadline{$\mu_{\text{low}}$} &  \tableheadline{$\bar{e}_{\text{low}}$} & \tableheadline{$\bar{s}_{\text{low}}$} & \tableheadline{$\mu_{\text{high}}$} & \tableheadline{$\bar{e}_{\text{high}}$} & \tableheadline{$\bar{s}_{\text{high}}$}  \\ \midrule
		Polyester    & 0   & 0.95 cm  & 1.20 cm   & 1 & 0.84 cm & 1.03 cm   \\ 
		Wool         & 0   & 0.58 cm  & 0.73 cm   & 2 & 0.52 cm & 0.75 cm   \\ 
		Stiff-cotton & 0   & 0.60 cm  & 0.86 cm   & 2 & 0.58 cm & 0.77 cm   \\
		Denim        & 0   & 0.77 cm  & 1.11 cm   & 1.6 & 0.61 cm & 0.80 cm   \\  
		\bottomrule
	\end{tabularx}
	\caption{\label{tabla_fricciones}Optimal values of the friction coefficients along with the mean absolute error and spatial standard deviation for the low and high friction scenario}
	
\end{table}

\begin{figure}[htb!]
	\centering
	\includegraphics[width=0.85\linewidth]{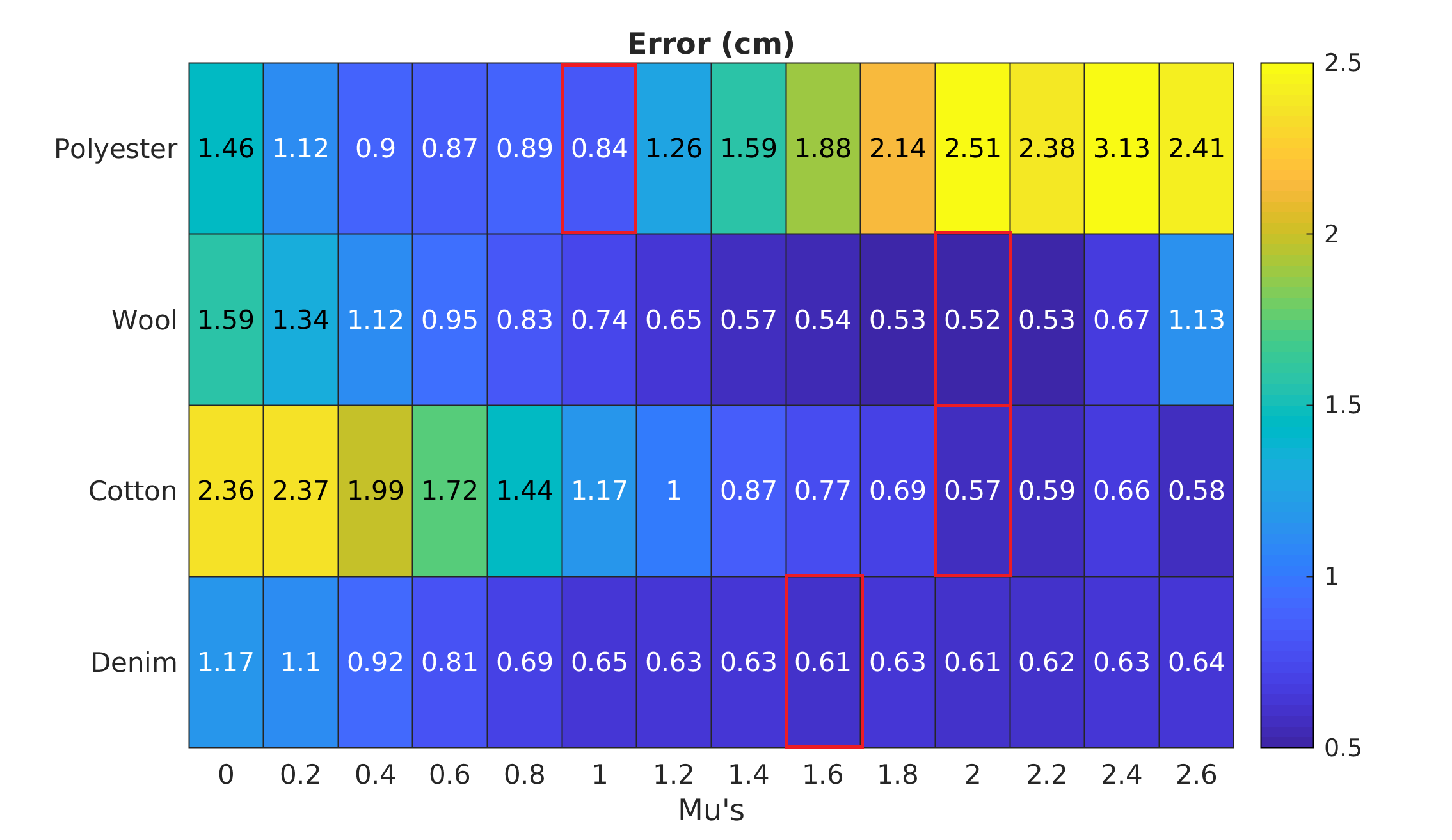}
	\caption{\label{sensi_friccion}Sensitivity analysis for the high friction case, i.e. we vary the value of $\mu$ and compute the absolute error (\ref{error_rel}) for the four A2 fabrics. In red we encircle the error found with the optimal parameter of $\mu$.}
\end{figure}

In order to understand how friction influences the dynamics of the textiles we perform a sensitivity analysis for the high friction case, i.e. we vary the value of $\mu$ (keeping all the other parameters fixed), and compute the mean of the absolute error (\ref{error_rel}). The results can be seen in the heat-map depicted in Figure \ref{sensi_friccion}. Notice that in general, the model is quite stable with respect to the optimal friction value.

\subsection{Hitting scenario}

\begin{figure}[htb!]
	\centering
	\includegraphics[width=1\linewidth]{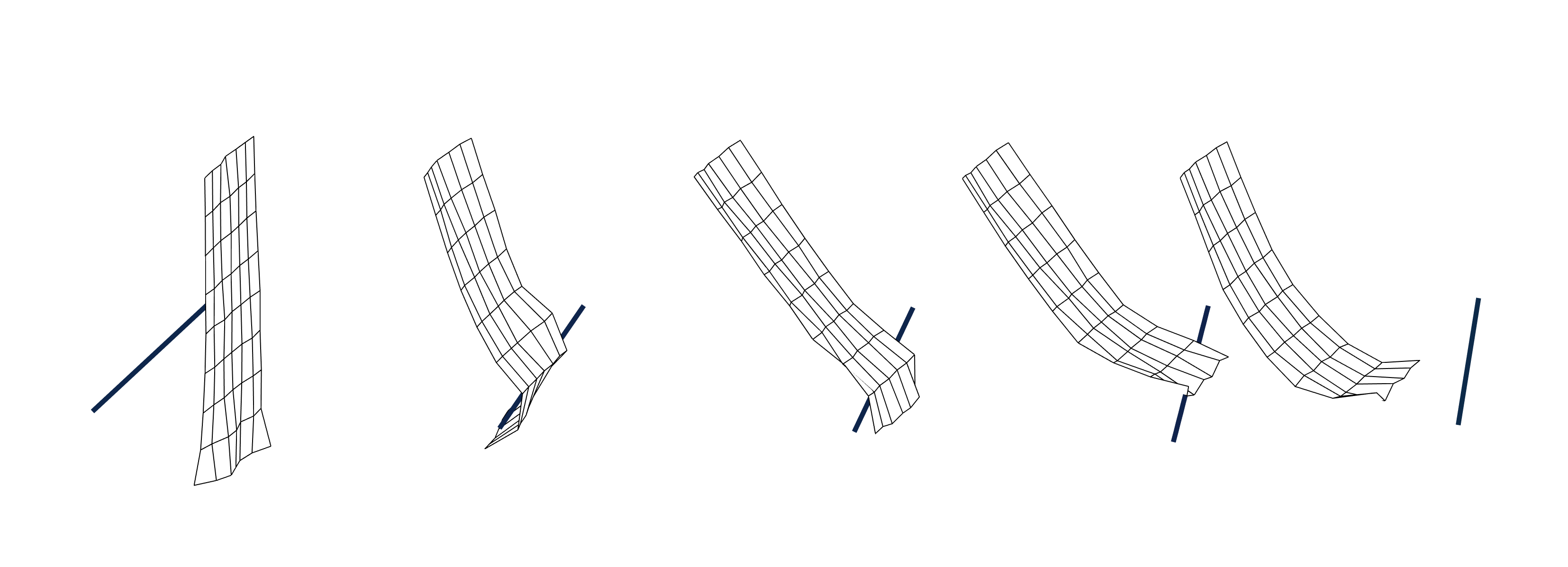}
	\caption{\label{manta_palo}Long-stick hits sequence (left to right): the cloth is held by its two upper corners and then is hit repeatedly with the cylindrical stick. The hits are aimed at different locations with varied intensities.}
\end{figure}

In this second scenario, the fabrics are held suspended in the air (with the long sides perpendicular to the floor) and hit repeatedly with a long stick. The hits are aimed at various locations of the cloth with varied strengths and speeds (see Figure \ref{manta_palo}). In order to simulate the hits, the stick is subdivided into small edges and we employ a procedure similar to the one used for self-collisions of the cloth in the case of an edge-edge collision (see Section \ref{sec_autocols}). 

Let us denote by $\{a_1(t),\dots, a_m(t)\}$ the endpoints of the edges of the stick. Then, for every iteration $j$ of the iterative process (\ref{prob_qua}), like we do with self-collisions (see Section \ref{autocol_iter}), we must check if a collision occurred during the motion between the stick-edges 

\begin{equation*}
	\{a_1(t_n),\dots, a_m(t_n)\}\rightarrow_{dt}\{a_1(t_{n+1}),\dots, a_m(t_{n+1})\}
\end{equation*}
and the edges of our triangulated cloth $\boldsymbol{\varphi}^{n}\rightarrow_{dt}\boldsymbol{\varphi}_{j}$. This means that for every detected collision, in the next iteration $j+1$ of the sequence of quadratic problems (\ref{prob_qua}) we must add a constraint of the form:  

\begin{equation*}
	H(\boldsymbol{\varphi}_{j+1}) = \langle \pi_{\alpha}(a_1,a_2) - \pi_{\beta}(x_1,x_2) ,\nu\rangle \geq 0,
\end{equation*}
where $a_1,a_2$ are the two endpoints of the corresponding edge of the stick, $x_1,x_2$ are likewise the two endpoints of the cloth's edge, $\pi_{\alpha'}(a_1,a_2) = (1-\alpha')a_1 +\alpha' a_2$ and $\pi_{\beta'}(x_1,x_2) = (1-\beta')x_3 +\beta' x_4$ are the closest points between the two segments and $\nu$ is the normal vector to both edges. The values $\nu,\alpha',\beta'$ are assumed to be constant in time, and are computed in the case of the cloth with the positions of the segments defined by $\boldsymbol{\varphi}_{j}$ and for the stick at time $t_{n+1}$. The normal vector $\nu$ is oriented such that $H(\boldsymbol{\varphi}^{n})\geq 0$. 

\medskip

On the other hand, the real long stick has a length of 75 cm and a diameter of 1.5 cm. Two markers with a diameter of 1.5 cm are put at both ends of the stick to record its trajectory. 

\remark We consider the stick's thickness by imposing $H(\boldsymbol{\varphi})\geq\tau_0$, where $\tau_0 = 0.75 \text{cm}$ is the radius of the stick. Moreover, this thickness is taken into account in the detection process (see Section \ref{grosor}). 

\medskip

The stick is made of polished plastic and hence we consider friction between the cloth and the stick to be negligible (moreover, since the cloth is held firmly by the two upper corners the small amount of friction that could exist is always overcome by the stick). Each textile is hit four times with recordings varying between 12 and 18 seconds (as usual with a frame every $dt = 0.01$ seconds). On top of fitting as usual the damping parameter $\alpha$ and the (virtual) gravitational mass $\delta$, the goal of this scenario is to assess the realism of our collision algorithm when modeling the hits.

\remark By the nature of this collision experiment, some movements of the textiles are very abrupt and therefore, as mentioned before, the markers disappear some of the time. This problem is also present in the recording of the trajectory of the stick; since the stick is rigid, to solve this issue we have found that it is enough to interpolate linearly its missing positions. 

\medskip

In this experiment, we also study the performance of the \textbf{active-set} collision algorithm described in Section  \ref{sec_active-set}. We compare it with a standard \textbf{interior-point} algorithm implemented to solve large sparse quadratic problems (e.g. as implemented in the MATLAB function \texttt{quadprog}, see Section 16.7 of \citep{Nocedal1999NumericalO}). We tried to compare our novel algorithm with a standard active-set method for quadratic optimization, but we were unable to find a working implementation in any programming language for large and sparse problems. All comparisons are performed using an Intel Core i7-8700K with 12 cores of 3.70 GHz. Since the four recordings have different durations, we compute the quotient

\begin{equation} \label{cociente_times}
	q = \frac{T_{\textbf{sim}}}{T_{\textbf{rec}}},
\end{equation}  
where $T_{\textbf{rec}}$ is the duration of the recording and $T_{\textbf{sim}}$ is the amount of time it takes to simulate it. Hence $q \approx 1$ would mean that the simulations work on real-time, $q \approx 0.5$ means that they are twice as fast, etc. In Table \ref{tabla_palo} we can see the value of the absolute error and standard deviation with the optimal value of the parameters $\alpha$ and $\delta$ (not shown).

\smallskip

\begin{table}[htb]
	\centering
	\begin{tabularx}{0.7\textwidth}{lcccc} \toprule
		\tableheadline{Material} & \tableheadline{$\bar{e}$} & \tableheadline{$\bar{s}$}  & \tableheadline{Active-set} & \tableheadline{Interior-point} \\ \midrule
		Polyester    & 1.44 cm   & 2.13 cm  & 0.456  & 1.344        \\ 
		Wool         & 1.39 cm   & 2.23 cm  & 0.437  & 1.298          \\ 
		Denim        & 0.98 cm   & 1.86 cm  & 0.425  & 1.235           \\ 
		Stiff-cotton & 1.07 cm   & 1.85 cm  & 0.510  & 1.576         \\ 
		\bottomrule
	\end{tabularx}
	\caption{\label{tabla_palo}Mean absolute error and spatial standard deviation with the optimal value of the parameters $\alpha$ and $\delta$ (not shown). In the two last columns, we display the quotient (\ref{cociente_times}), for our active-set collision algorithm and a standard interior-point method.}
	
\end{table}

\begin{figure}[htb!]
	\centering
	\includegraphics[width=0.8\linewidth]{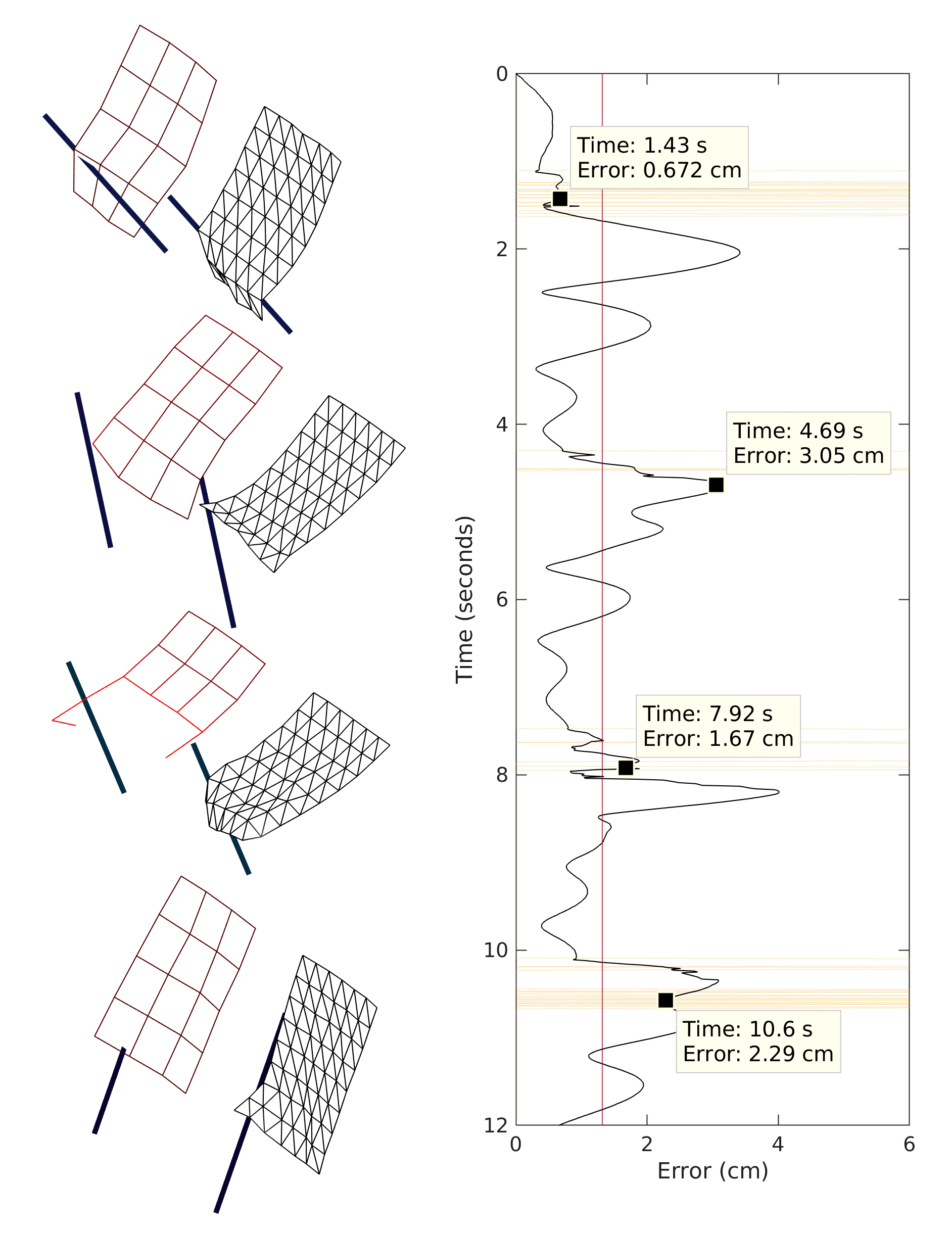}
	\caption{\label{palo_algodon} Four frames comparing the recorded hitting of A2 stiff-cotton (left) with its inextensible simulation (right); its average error being 1.07 cm. On the right, we show a full plot (vertically) of the absolute error and with yellow lines, we highlight the moments in which the stick is in contact with the cloth.}
\end{figure}

For a visual comparison of the results, together with a plot of how the absolute error varies with time for the four textiles, see Figure \ref{palo_algodon} (stiff-cotton), the graphical abstract at the beginning of the paper (polyester) and \url{https://youtu.be/U7-p_1E09L8} (for all four materials, including denim and wool). With yellow lines, we highlight the moments in which the stick is in contact with the cloth (during the simulations). Notice that it is precisely in those instants where more missing data is found. In the figures we see clearly that the error concentrates after the hit and not during it, showing that the collision model is very realistic but afterwards the aerodynamics become dominant and the errors increase. Overall the fitting is quite good, with errors slightly bigger than those found in Section \ref{sec_tablecloth}. Finally, our active-set algorithm is found to be almost 3 times faster than a standard interior-point method, with simulation going faster than real-time for a $7\times 9$ mesh (see Table \ref{tabla_palo}). 

\section{Conclusions and further work}

In this work, we delved into the problem of modeling collisions, including the response, for inextensible cloth. We explained how to incorporate contacts with an object (using Signorini's conditions), self-collisions, and Coloumb friction into the equations of motion in such a way that the model integrates all constraints (inextensibility and contacts) and friction forces at the same time without any decoupling. We developed a novel numerical discretization of contact and friction forces that can be seen as a natural extension of the \textit{fast projection algorithm} presented in \cite{Goldenthal:2007:ESI} in order to include inextensibility, contacts, and friction in a single pass. This discretization led naturally to a sequence of quadratic problems with inequality and equality constraints. We presented a novel \textit{active-set} method tailored to our problem which takes into account past active constraints to accelerate the resolution of unresolved contacts. The main advantage of this new algorithm with respect to standard \textit{active-set} methods is its ability to start from any point and not necessarily from a feasible one. Moreover, we showed with different simulations that our model of friction is effective in static and dynamic settings, that collisions with sharp objects can be easily included and that complex folding sequences of cloth with non-trivial topologies (a pair of shorts) can be performed. 

\medskip

Finally, with the aid of a \textit{Motion Capture System}, we embarked ourselves on the empirical validation of the developed model. We validated two different but related aspects of the collision model: its ability to simulate properly friction and to model the dynamics of fast and strong hits with a rigid object. We found the optimal friction parameters for both a high and a low friction case (see Table \ref{tabla_fricciones}), with absolute errors under 1 cm for four DIN A2 textiles. We showed that the simulations are very stable with respect to friction by performing a sensitivity analysis. Furthermore, using only two parameters, we were able to model the most challenging scenario of this paper: four size DIN A2 cloths were held by their two upper corners and then hit repeatedly at different locations and varied intensities with a long stick. The average errors are all around 1 cm and we were able to properly simulate the hits, appearing the biggest errors not during the hits but just after because of aerodynamic effects. The simulations on a desktop computer are two times faster than real-time (for the hitting scenario with a $7\times 9$ mesh), being our novel active-set solver three times faster than a standard interior-point method using the same mesh resolution.

\medskip

The work presented here essentially completes the task initiated by the
authors in \cite{COLTRARO2022}: the development and implementation of a mechanical model for cloth able to simulate common tasks in a human environment, such as folding, in real-time or faster, in a way that is faithful to the real behavior of cloth, with a margin of error of order 1 cm for a typical garment size. Realism, rather than \textit{spectacularity} of the model is crucial for its usefulness in the training of Machine Learning algorithms, such as Neural Networks, to physically perform these tasks using robotic arms. The extensive work reported here for its validation has been undertaken to make our model capable of replacing actual physical manipulations of cloth with a robot and thus speeding up the training of such algorithms.

\medskip

Because of this reason, the natural continuation of this work will be
the application of Control Theory to the robotic control of cloth
manipulation, both through a classical, deterministic approach and also using Deep and Reinforcement Learning methods. As a first step in this direction, the authors are currently integrating an implementation of the here presented model in a Virtual Reality environment to make data collection faster and easier for this application to Robotics.

\section{Acknowledgments}
This work was developed in the context of the project CLOTHILDE (``CLOTH manIpulation Learning from DEmonstrations") which has received funding from the European Research Council (ERC) under the European Union's Horizon 2020 research and innovation programme (grant agreement No. 741930).
M. Alberich-Carrami\~nana is also with the Barcelona Graduate School of Mathematics (BGSMath) and the Institut de Matem\`atiques de la UPC-BarcelonaTech (IMTech), and she is partially supported by the grant PID2019-103849GB-I00 funded by MCIN/ AEI /10.13039/501100011033.

\bibliographystyle{elsarticle-num-names} 
\bibliography{Bibliography}
\end{document}